\documentclass{article}
\usepackage[utf8]{inputenc}
\usepackage{graphicx}
\usepackage{amsmath}
\usepackage[authoryear]{natbib} 
\usepackage{hyperref}
\usepackage{etoolbox}
\makeatletter
\patchcmd{\NAT@citex}
  {\@citea\NAT@hyper@{\NAT@nmfmt{\NAT@fullname}~\NAT@date}}
  {\@citea\NAT@nmfmt{\NAT@fullname}~\NAT@hyper@{\NAT@date}}
  {}{}
\makeatother

\usepackage{float}
\usepackage{tabularx} 
\usepackage{booktabs} 
\usepackage{csquotes}
\usepackage[normalem]{ulem} 
\usepackage[none]{hyphenat}
\usepackage{placeins} 
\usepackage{longtable}  
\usepackage{pdflscape}
\usepackage{caption}
\usepackage[normalem]{ulem}
\usepackage{adjustbox} 
\usepackage{xcolor}
\usepackage{fancyhdr}
\usepackage{colortbl}    
\usepackage{siunitx} 
\usepackage{authblk}
\usepackage[margin=1in]{geometry}
\usepackage{booktabs}  
\usepackage{multicol} 
\usepackage{subcaption}
\usepackage{chngcntr}
\definecolor{lightred}{HTML}{FDE0E0}
\definecolor{lightgreen}{HTML}{E0FDE0}
\title{Lost in Cultural Translation: Do LLMs Struggle with Math Across Cultural Contexts?}

\author[1]{Aabid Karim\textsuperscript{*}\thanks{\texttt{abed.karim@55mv.co}}}
\author[2]{Abdul Karim\textsuperscript{*}\thanks{\texttt{abdulkarim@microsoft.com}}}
\author[1]{Bhoomika Lohana\textsuperscript{*}\thanks{\texttt{bhoomikalohana70@gmail.com}}}
\author[1]{Matt Keon\thanks{\texttt{mattk@55mv.co}}}
\author[3]{Jaswinder Singh\thanks{\texttt{jaswinder@millcrest.com.au}}}
\author[4]{Abdul Sattar\thanks{\texttt{a.sattar@griffith.edu.au}}}

\affil[1]{55mV Research Lab}
\affil[2]{Microsoft}
\affil[3]{Millcrest Technology}
\affil[4]{Griffith University}

\begin{document}

\maketitle
\setcounter{footnote}{0}
\renewcommand{\thefootnote}{}  
\phantomsection 
\footnotetext{\textsuperscript{*}These authors contributed equally to this work.}
\renewcommand{\thefootnote}{\arabic{footnote}}  

\begin{abstract}
We demonstrate that large language models' (LLMs) mathematical reasoning is culturally sensitive: testing 14 models from Anthropic, OpenAI, Google, Meta, DeepSeek, Mistral, and Microsoft across six culturally adapted variants of the GSM8K benchmark, we find accuracy drops ranging from 0.3\% (Claude 3.5 Sonnet) to 5.9\% (LLaMA 3.1-8B) when math problems are embedded in unfamiliar cultural contexts, even when the underlying mathematical logic remains unchanged. These statistically significant performance reductions (p $<$ 0.01, confirmed through McNemar tests) reveal that mathematical reasoning in LLMs is not culturally neutral.

To create these variants for Haiti, Moldova, Pakistan, Solomon Islands, Somalia, and Suriname, we systematically replaced cultural entities (names, foods, places, etc.)  in 1,198 GSM8K questions while preserving all mathematical operations and numerical values. Validation through 130 randomly sampled questions with two independent human annotators (73.8\% and 79.2\% accuracy) confirmed entity recognition quality, with discrepancies manually corrected. Our quantitative error analysis of 18,887 instances reveals that while explicitly cultural errors (currency: 4.0\%, relationships: 4.5\%) represent a minority, cultural adaptation affects broader  reasoning patterns, with mathematical reasoning errors comprising 54.7\% and calculation errors 34.5\% of failures.

Interestingly, cultural familiarity can enhance performance: Mistral Saba, despite lacking explicit mathematical training, outperforms some larger models on Pakistan-adapted problems due to Middle Eastern and South Asian training data exposure. Our analysis suggests multiple mechanisms contribute to performance degradation, including tokenization differences, unfamiliar cultural references disrupting reasoning pathways, and training data biases toward Western contexts. This study underscores the need for more diverse training data to ensure robust LLM performance across global contexts. 
\end{abstract}

\section{Introduction}
Large Language Models (LLMs) have transformed AI, taking on tasks that once seemed to require human intuition. These models have shown remarkable progress in the field of natural language processing, problem solving, question answering, and computer vision tasks \citep{li2025visual}, \citep{chu2024visionllama}, \citep{gunter2024apple}, \citep{team2023gemini}. Their ability to interpret and code complex mathematical reasoning has gained a tremendous amount of attention from the research community \citep{mirzadeh2024gsm}. 

However, beneath their impressive capabilities lies a crucial question: Do LLMs truly understand the world's cultural diversity, or reflect simply the cultural limitation embedded in their training data? More specifically, do they retain their mathematical reasoning when presented with culturally adapted math word problems? How does cultural diversity—or lack thereof—in their training data impact their mathematical reasoning abilities? Studies have consistently highlighted cultural, gender, and sociopolitical biases in LLMs, particularly their Western-centric learning, which affects fairness and adaptability in different linguistic and cultural contexts \citep{ramesh2023fairness, ramezani2023knowledgeculturalmoralnorms}. \citet{hovy2016social} argue that such biases arise because NLP systems reflect the social and cultural characteristics of their training data, often disadvantaging underrepresented populations. Prior work has further shown that surface-level changes in representation, such as names or character roles, can systematically affect model outputs even when the underlying task structure remains constant \citep{lucy-bamman-2021-gender}, and that models exhibit culturally skewed tendencies in multilingual settings \citep{naous2024havingbeerprayermeasuring}. 

Beyond cultural bias, LLMs exhibit fundamental limitations in formal reasoning. Research suggests that LLMs do not engage in formal reasoning \citep{valmeekam2023planbenchextensiblebenchmarkevaluating} but instead rely on probabilistic pattern-matching to generate outputs based on training data rather than a true conceptual understanding of symbols or ideas \citep{boixadsera2024transformersreasonabstractsymbols}. Studies have shown that LLMs struggle with formal logical reasoning, as their outputs are highly sensitive to individual token changes \citep{jiang2024peektokenbiaslarge}. A single transformer layer functions similarly to a one-nearest neighbor algorithm \citep{li2024onelayertransformerprovablylearns}, meaning the model's reasoning process is heavily influenced by the closest matching examples in its training data. The relationship between training frequency and test performance further supports this idea \citep{razeghi2022impactpretrainingtermfrequencies}, demonstrating that LLMs rely more on statistical associations than genuine logical inference. 

These reasoning limitations intersect with cultural context in a specific way. LLMs are highly sensitive to input tokens, and even slight changes in their tokenization process can alter their reasoning \citep{grattafiori2024llama}. If they have not been exposed to diverse and underrepresented cultural norms and contexts during training, they may tokenize culturally specific prompts differently, potentially leading to shifts in reasoning and varied responses. Furthermore, as the number of tokens increases or tasks become more complex, the probability of accurately predicting tokens decreases \citep{shi2023large}.

To address these questions, we use the GSM8K dataset \citep{cobbe2021training}, a widely recognized benchmark to evaluate the mathematical reasoning capabilities of LLM \citep{mirzadeh2024gsm}. However, GSM8K has two key limitations. First, given its extensive use, there is a high probability that LLMs have encountered this dataset during training, making it less reliable for assessing their reasoning abilities. Second, it consists of a single set of math problems, lacking any cultural diversity. To overcome these limitations, we modify the GSM8K test set by introducing culturally adapted versions of each question. Specifically, we synthesize six culturally diverse variants from the original GSM8K test set , one for each continent, while preserving the original mathematical logic and numerical values. The only changes involve replacing culturally specific entities such as names, food items, and contextual references with those relevant to each target culture. We then evaluate 14 LLMs, varying in size and release period, in these culturally adapted versions of GSM8K test set. 

A detailed comparison of this work with prior related research is provided in Section 2. The remainder of the paper is structured as follows. Section 3 details the dataset creation process, Section 4 discusses our culture (country) selection process, Section 5 shows detailed performance analysis of LLMs, and Sections 6 presents the conclusion.

\section{Related Work}
A growing body of research has documented that NLP systems reflect the social and cultural characteristics of their training data, often disadvantaging underrepresented populations \citep{hovy2016social}, with fairness research remaining concentrated predominantly on English and high-resource languages \citep{ramesh2023fairness, ramezani2023knowledgeculturalmoralnorms}. Two lines of work are particularly relevant to our study. \citet{lucy-bamman-2021-gender} demonstrate that surface-level changes in representation, specifically character gender in narrative prompts, systematically affect GPT-3 outputs in topics and power dynamics, even when the task structure remains constant. \citet{naous2024havingbeerprayermeasuring} introduce CAMeL, a dataset contrasting Arab and Western cultural entities, showing that both multilingual and Arabic monolingual LMs exhibit significant bias towards Western entities across tasks including story generation, NER, and sentiment analysis. While these studies establish that cultural context shapes model outputs, both evaluate open-ended tasks without objectively verifiable correct answers, and neither examines mathematical reasoning or spans multiple underrepresented cultures across continents. Our work addresses these gaps by evaluating 14 LLMs on culturally adapted math problems with ground truth answers across six underrepresented cultural contexts spanning six continents. 

A related line of research examines the sensitivity of LLM reasoning to surface-level input changes. \citet{jiang2024peektokenbiaslarge} introduce a hypothesis-testing framework demonstrating that LLMs rely on token bias rather than genuine reasoning, by systematically altering tokens such as names and quantifiers in logical fallacy problems while preserving the underlying logic, they show that performance drops significantly, confirmed through McNemar testing. Similarly, \citet{shi2023large} introduce GSM-IC, a variant of GSM8K augmented with artificially inserted irrelevant sentences, finding that model performance dramatically decreases when such distractors are present. Both studies establish that LLMs are sensitive to surface-level contextual changes in reasoning tasks. However, the perturbations in both cases are arbitrary or artificial, designed specifically to expose brittleness, and neither study introduces a cultural dimension. Our central finding is that cultural unfamiliarity disrupts mathematical reasoning even when the underlying problem difficulty and numerical logic remain entirely unchanged,  suggesting that model performance on mathematical reasoning tasks may be sensitive to the cultural framing of problems, rather than purely to their mathematical structure.

The GSM8K benchmark introduced by \citet{cobbe2021training} has become the standard for evaluating mathematical reasoning in LLMs, comprising grade-school level math word problems requiring multi-step arithmetic reasoning. While LLM performance on GSM8K has improved substantially, \citet{mirzadeh2024gsm} demonstrate through their GSM-Symbolic benchmark that this improvement may not reflect genuine reasoning advances,  by substituting names and numerical values using symbolic templates, they show that model performance exhibits noticeable variance across different instantiations of the same problem. Our work is directly inspired by this approach of controlled substitution while preserving mathematical logic. However, where GSM-Symbolic varies mathematical surface features within a single Western English cultural context, we vary cultural entities across six real-world cultural contexts spanning six continents. Importantly, our findings reveal that this relationship is bidirectional:  while unfamiliar cultural contexts degrade mathematical reasoning performance, cultural familiarity can enhance it, a pattern observed when models with regional training data exposure outperform larger models on culturally familiar variants. This bidirectional effect has no analogue in prior benchmark work. Additionally, research on tokenization fairness suggests that culturally specific vocabulary may be processed less efficiently by models trained predominantly on English text \citep{petrov2023language}, \citep{dang2024tokenization}, representing one potential contributing factor to the performance differences we observe.

\section{Dataset Creation}

\begin{figure}[H]
    \centering
    \includegraphics[width=1\textwidth]{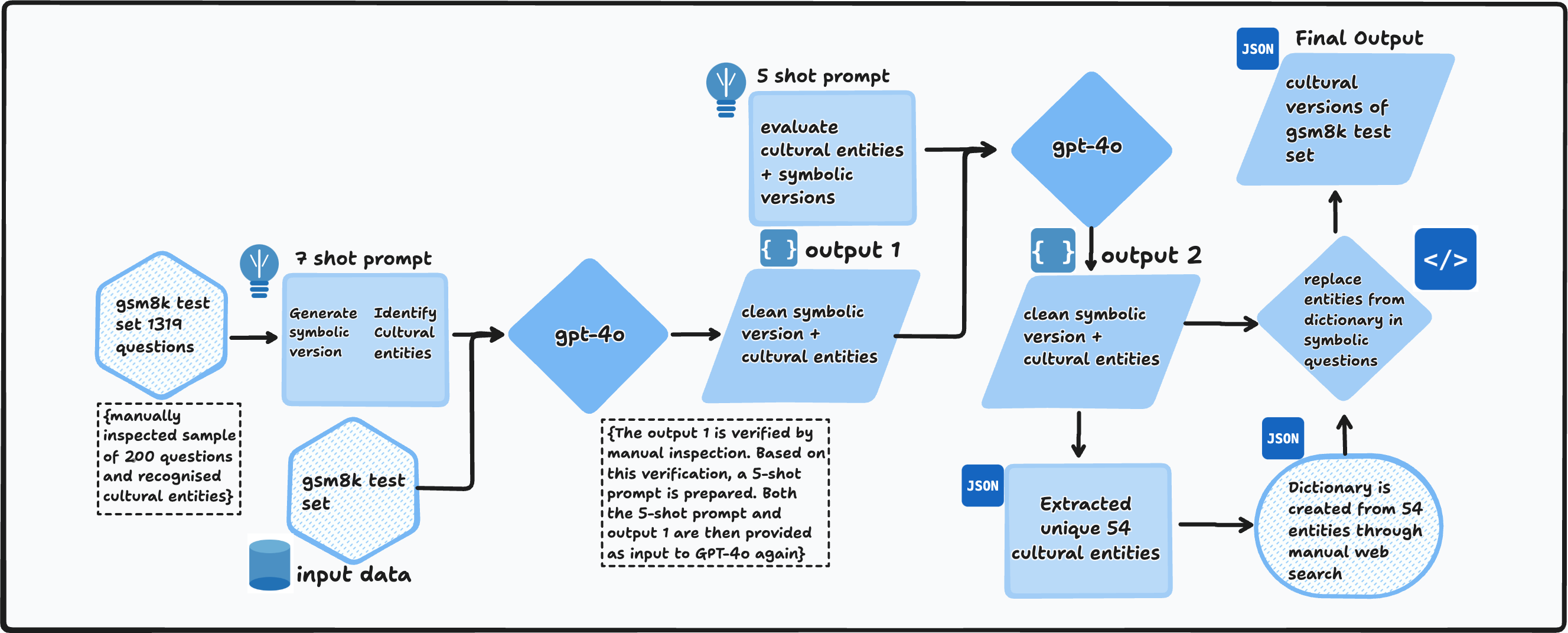}
    \caption{Cultural Datasets Creation Flow}
    \label{fig:main_flow}
\end{figure}

Our dataset creation process and structure are strongly inspired by the GSM symbolic dataset introduced by \citep{mirzadeh2024gsm}. They used the GSM8K test dataset to generate symbolic variants of questions by replacing names with placeholders. This approach enables variations in names while preserving the core mathematical logic, reasoning, and numerical structure of GSM8K. Additionally in their research, numerical values are also adjusted to introduce further complexity to the questions in the GSM8K test set.
Similarly, we used the GSM8K test dataset, converting a total of 1,319 questions into symbolic duplicates. However, our approach extends beyond just name substitutions. We systematically identify various cultural entities present in the questions, including person names, food items, clothing, city names, school subjects, common sports, etc. Each identified entity is carefully examined and replaced, ensuring that the questions reflect different cultural contexts without altering their mathematical logic. Importantly, we maintain all the numerical values and the original structure of the questions, making targeted adjustments only to reflect cultural variations. The general data set creation process is shown in Figure~\ref{fig:main_flow}.

\subsection{Cultural Entities Recognition}
Initially, we select a representative sample of 200 questions from the 1,319 questions in the GSM8K dataset and manually identify cultural entities through a detailed human evaluation. Subsequently, we manually create symbolic versions of seven randomly chosen questions from this subset, replacing the identified cultural entities with accurate placeholders. These manually prepared symbolic questions form the basis of a 7-shot prompt constructed for GPT-4o provided in Figure~\ref{fig:5} (Appendix A).

Using the established 7-shot prompt, we employ GPT-4o to systematically recognize cultural entities throughout the entire test dataset, processing the questions in batches of 100. GPT-4o outputs each batch in a structured format, clearly delineating the original questions, identified entities, and symbolic versions, as shown in Figure~\ref{fig:sub1}. From the GPT-4o output, we extract unique cultural entities identified in all batches. Furthermore, we find that 121 questions did not contain any identifiable cultural entities, reducing the total number of culturally adaptable questions to 1,198.

To validate the quality of GPT-4o's cultural entity recognition, we conduct a systematic evaluation on 130 randomly sampled questions, representing approximately 11\% of the culturally adaptable dataset. Two independent human annotators manually review each question to verify whether GPT-4o correctly identified all cultural entities without omissions or errors. The annotators achieve question-level accuracy scores of 73.8\% and 79.2\%, respectively, indicating that GPT-4o successfully identified all entities in approximately three out of four questions. We adopted question-level accuracy as our validation metric, as our dataset required every question to have all cultural entities correctly identified,  a single missed cultural entity breaks the logical consistency of the entire question, making it unusable regardless of how many other entities were correctly identified. All identified discrepancies from the validation sample were manually reviewed and corrected. Importantly, the correction process was not limited to the sampled questions,  informed by the error patterns identified during human validation, we conducted multiple iterative correction cycles across the full dataset, combining GPT-4o automated evaluation with human inspection drawing on our cultural familiarity with the target regions. A detailed description of the validation methodology and results is provided in Appendix A Section~\ref{Validation of Cultural Entity Recognition Quality} as well as the broader multi-layer correction process applied to the full dataset.  The iterative correction process for placeholder naming inconsistencies is described in Appendix A Section~\ref{appendix:entity-correction}.

Finally, we obtain a clean and standardized symbolic version of each question, with only identified cultural entities and placeholders at their specific locations, as well as a clean set of 54 cultural entity types. An example question can be seen in Figure~\ref{fig:sub1}, and the full list of cultural entities is included in Table~\ref{tab:2} (Appendix A). 

\subsection{Dictionary Creation}
After obtaining a clean symbolic version with \textbf{54 cultural entity types}, we create a dictionary for each culture. The process of selecting different cultures (countries) is explained in Section ~\ref{section:3}. This dictionary is primarily constructed using manual web searches, Wikipedia, country-specific websites, and blogs to gather information about each cultural entity. For instance, if a cultural entity is \textit{`Person name'}, we search the Web, Wikipedia, and various country-specific sources to collect the most common names associated with that culture. Similarly, for the entity \textit{`food items'}, we look for traditional foods specific to that culture and so on for the rest of the entities. 

In the dictionary, each cultural entity serves as a key (e.g., ``Person name"), and the corresponding values are the most common names or items related to that entity. The number of values in the dictionary is directly proportional to how frequently that entity appears in the dataset. For example:
\begin{itemize}
    \item \textbf{Person name} occurs the most in our dataset, so we have the largest number of values (names) in the dictionary for the person name entity.
    \item \textbf{Food item} is the second most frequent entity, so it has the next highest number of values, and so on accordingly, for the rest of the entities.
\end{itemize}

The scraping process to fill the placeholder with various cultural entities is performed manually, allowing us to inspect and validate each entry to ensure accuracy. This also gives us the opportunity to review our list of identified cultural entities again, so we can remove any discrepancies. Through this process, we build and populate our dictionary for each culture. The values in the dictionary are then used to replace the placeholders in the symbolic version of our dataset. A snapshot of a dictionary is given in Figure~\ref{fig:7} (Appendix A~).

\subsection{Mapping Rules}
\begin{figure}[h]
    \centering
    \begin{subfigure}[b]{0.6\textwidth}
        \includegraphics[width=\textwidth]{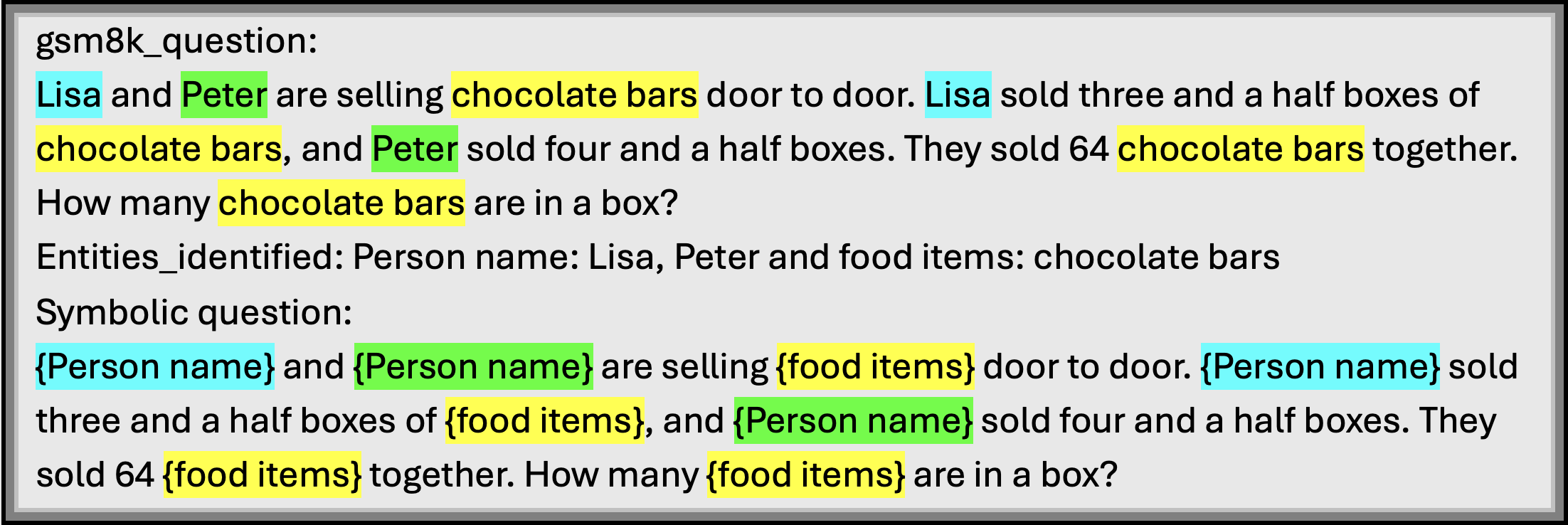}
        \caption{Symbolic version of an original sample question from GSM8K test dataset}
        \label{fig:sub1}
    \end{subfigure}
    \hfill
    \begin{subfigure}[b]{0.6\textwidth}
        \includegraphics[width=\textwidth]{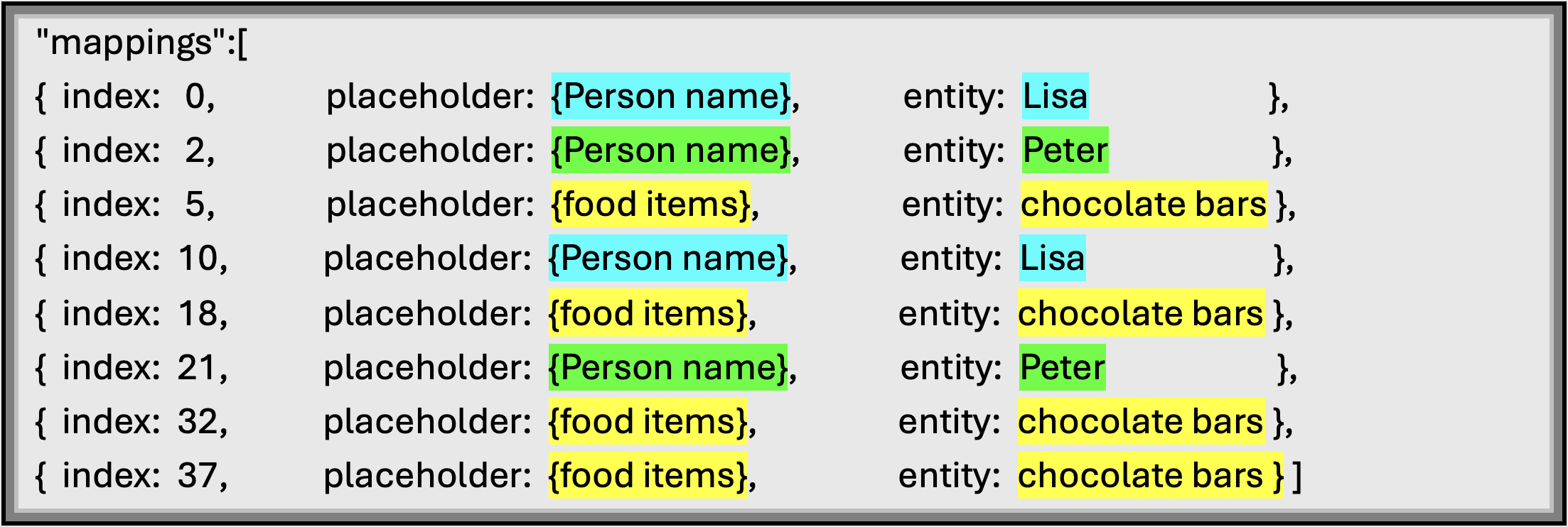}
        \caption{Mapping rules for the sample question from GSM8K test dataset}
        \label{fig:sub2}
    \end{subfigure}
    \hfill
    \begin{subfigure}[b]{0.6\textwidth}
        \includegraphics[width=\textwidth]{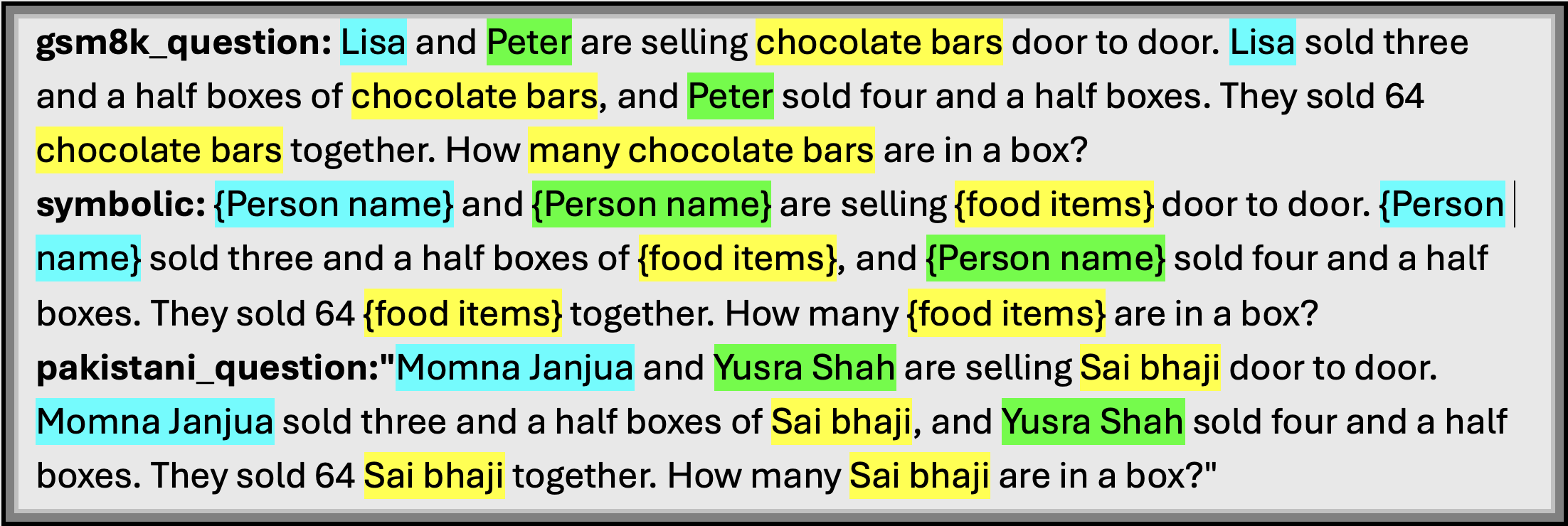}
        \caption{Original GSM8K test set sample question, its symbolic version and its cultural variant after
        replacement}
        \label{fig:sub3}
    \end{subfigure}
    \caption{}
    \label{fig:main}
\end{figure}

One of the final challenges in our dataset creation is ensuring that the mathematical logic of the cultural dataset questions remains consistent with the original GSM8K test dataset, even after replacing placeholders with cultural entities from our dictionary. The objective is to precisely substitute only cultural entities in their designated placeholders while maintaining consistency for entities that appear multiple times within the same question; for example, if the same person name appears multiple times, it must be replaced with the same culturally adapted name at every occurrence. To achieve this, we create an indexing or mapping rule for each question, as illustrated in Figure~\ref{fig:sub2}, which establishes a one-to-one mapping between each placeholder instance and its corresponding original entity. This method removes randomness in entity replacement and ensures that the logical structure of every question remains unchanged across all entity types in our dictionary.

\subsection{Replacement}
Now, we have a clean dataset of symbolic questions, a set of well-defined mapping rules for each question, and a dictionary for each culture. Our dataset consists of 1,198 questions from the GSM8K test set, along with an exact symbolic version of those specific questions. To generate culturally adapted questions, we use a simple Python script to replace placeholders with cultural entities while strictly following the mapping rule for each question. The process works as follows:  

\begin{itemize}
    \item The mapping file is read simultaneously to locate placeholders in the tokenized question using predefined indices.  
    \item For each placeholder, its type (e.g., \texttt{\{Person name\}}) is identified, and a corresponding entity is selected from the cultural dictionary.  
    \item A tracking dictionary ensures that repeated placeholders in the same question receive the same entity throughout, maintaining logical consistency.  
    \item The updated tokens are then reconstructed into a complete question, ensuring that the structure, logic, and mathematical reasoning remain identical to the original GSM8K test question.  
\end{itemize}

This systematic replacement process ensures that our culturally adapted dataset remains logically consistent while introducing culturally relevant variations for evaluation. An example question can be seen in the Figure~\ref{fig:sub3}. 

\section{Country Selection}
\label{section:3}
For the initial selection of a country, we chose Pakistan, as the authors have first-hand cultural knowledge and lived experience in this region. This allowed us to ensure a more accurate and contextually relevant adaptation of the dataset. Since a deep understanding of cultural nuances is essential for meaningful modifications, selecting a country with which the authors are familiar provides a reliable foundation for this study. Beyond this, we established a structured selection criterion for additional countries. We aimed for broad representation by choosing one country from each continent (excluding Antarctica due to the absence of a permanent population) and prioritizing underrepresented and economically disadvantaged countries. These countries were chosen to reflect diverse cultural and socioeconomic contexts that may lack strong representation in AI and technology. By combining personal expertise with a systematic selection process, we strive to create a more comprehensive and meaningful evaluation of cultural biases in LLMs. We used three indicators given below to define underrepresented and economically disadvantaged countries. 

\begin{itemize}
    \item \textbf{Human Development Index (HDI):} Published by the United Nations Development Program (UNDP). It combines life expectancy, education, and per-capita income indicators.
    \item \textbf{Gross National Income (GNI) per capita:} Used by the World Bank to classify countries into low-income, lower-middle, upper-middle, and high-income categories.
    \item \textbf{Least Developed Countries (LDC):} A UN-designated list based on income per capita, human assets (health, education) and economic vulnerability.
\end{itemize}

We define a country as ``underrepresented and economically disadvantaged" if it is classified by the UN as a Least Developed Country (LDC) and appears in the bottom quartile of the Human Development Index (HDI) rankings for its region. To ensure that our research covers a broad range of `underrepresented' countries, one from each of the five populated continents (Africa, Europe, North America, South America and Oceania), we select based on the following criteria:
\begin{itemize}
    \item Countries are on the list of LDC of the United Nations for their region and appear in the bottom quartile of the Human Development Index (HDI) rankings.  
    \item In the absence of an LDC in that region, for example, Europe rarely has LDCs,  we select the lowest-ranked country on the Human Development Index as well as falling below a certain Gross National Income (GNI) index. 
\end{itemize}
Beyond economic metrics, we ensure that countries also reflect distinct languages, ethnicities, or cultural norms, as our study focuses on cultural biases.  Table~\ref{tab:country_dataset} contains the list of selected countries.

\begin{table}[H]
    \centering
    \begin{tabular}{|c|c|c|c|}
        \hline
        No & Country & Continent & Dataset \\
        \hline
        1 & Pakistan        & Asia           & PakGSM8K \\
        2 & Moldova         & Europe         & MolGSM8K \\
        3 & Somalia         & Africa         & SomGSM8K \\
        4 & Haiti           & North America  & HaiGSM8K \\
        5 & Suriname        & South America  & SurGSM8K \\
        6 & Solomon islands & Oceania        & SolIGSM8K \\
        \hline
    \end{tabular}
    \caption{Countries and Datasets}
    \label{tab:country_dataset}
\end{table}

\section{Results}
\subsection{Evaluation Methodology}
\label{4.1}
To assess the mathematical reasoning capabilities of LLMs in culturally adapted math problems, we systematically evaluate their accuracy across six cultural variants of the GSM8K test set. Each model is prompted with identical math problems, including the original GSM8K test dataset as well as its culturally modified versions for Haiti (HaiGSM8K) , Moldova (MolGSM8K) , Pakistan (PakGSM8K) , Solomon Islands (SolIGSM8K), Somalia (SomGSM8K), and Suriname (SurGSM8K). We select a diverse set of models from Anthropic, OpenAI, Google, Meta, DeepSeek, Mistral, and Microsoft based on both the size and release timeline to assess how mathematical reasoning evolves across different architectures. Our selection includes smaller models with relatively fewer parameters and larger models, allowing us to examine how scale impacts the performance of these models for our culturally adapted datasets.

To ensure a controlled evaluation, we keep all hyperparameters at their default values constant across models and use an identical prompting strategy. Specifically, no temperature parameter was explicitly set, ensuring that all models were evaluated under identical API default configurations rather than artificially tuned settings.  Each model attempts every question three times, generating three independent responses per dataset.  The prompt used for the evaluation is provided in Figure~\ref{fig:8} (Appendix B).

We evaluate accuracy by comparing model-generated answers with the ground truth from the GSM8K test set. Accuracy is defined as the number of correctly answered questions divided by the total number of questions. To ensure reliability, we adopt a strict consistency accuracy metric hereafter called strict accuracy: A question is considered correct only if all three generated responses exactly match the ground truth. If even one of the three responses is incorrect, the question is marked as incorrect. This approach mitigates random correct guesses and ensures that the accuracy reflects consistent performance rather than chance. 

To ensure robust accuracy estimates, we calculate 95\% Confidence Intervals (CIs) using the Wilson score interval, which provides more reliable estimates than the normal approximation, particularly when accuracy values are near the extremes, such as 0\% or 100\%. In our case, model accuracy tends to be high (typically around 95\%), with only a small proportion of questions answered incorrectly. Since accuracy is computed as a proportion of binary outcomes, each question is either fully correct or not to create a Bernoulli distribution. The Wilson method accounts for this distributional shape and corrects for the skew introduced near the boundary values. 

To determine whether models perform significantly different or not in cultural variants, we use McNemar’s test, a paired significance test that compares two datasets. We define our hypotheses as follows.
\begin{itemize}
    \item {Null Hypothesis (H\textsubscript{0})}: Cultural adaptation of the question does not affect the model’s accuracy. (The model is equally likely to be correct on GSM8K and its cultural variant.)
    \item {Alternate Hypothesis (H\textsubscript{1})}: The model is more likely to be correct on the original GSM8K question than on its culturally adapted version. (Cultural adaptation leads to lower accuracy.)
\end{itemize}

McNemar’s test counts: \textbf{b} $\rightarrow$ Cases where the model is correct on GSM8K but incorrect on the cultural variant. \textbf{c} $\rightarrow$ Cases where the model is incorrect on GSM8K but correct on the cultural variant.

\subsection{Performance}
\subsubsection{Model Accuracy Across Cultural Variants}
\begin{figure}[H]
    \centering
    \includegraphics[width=1\textwidth]{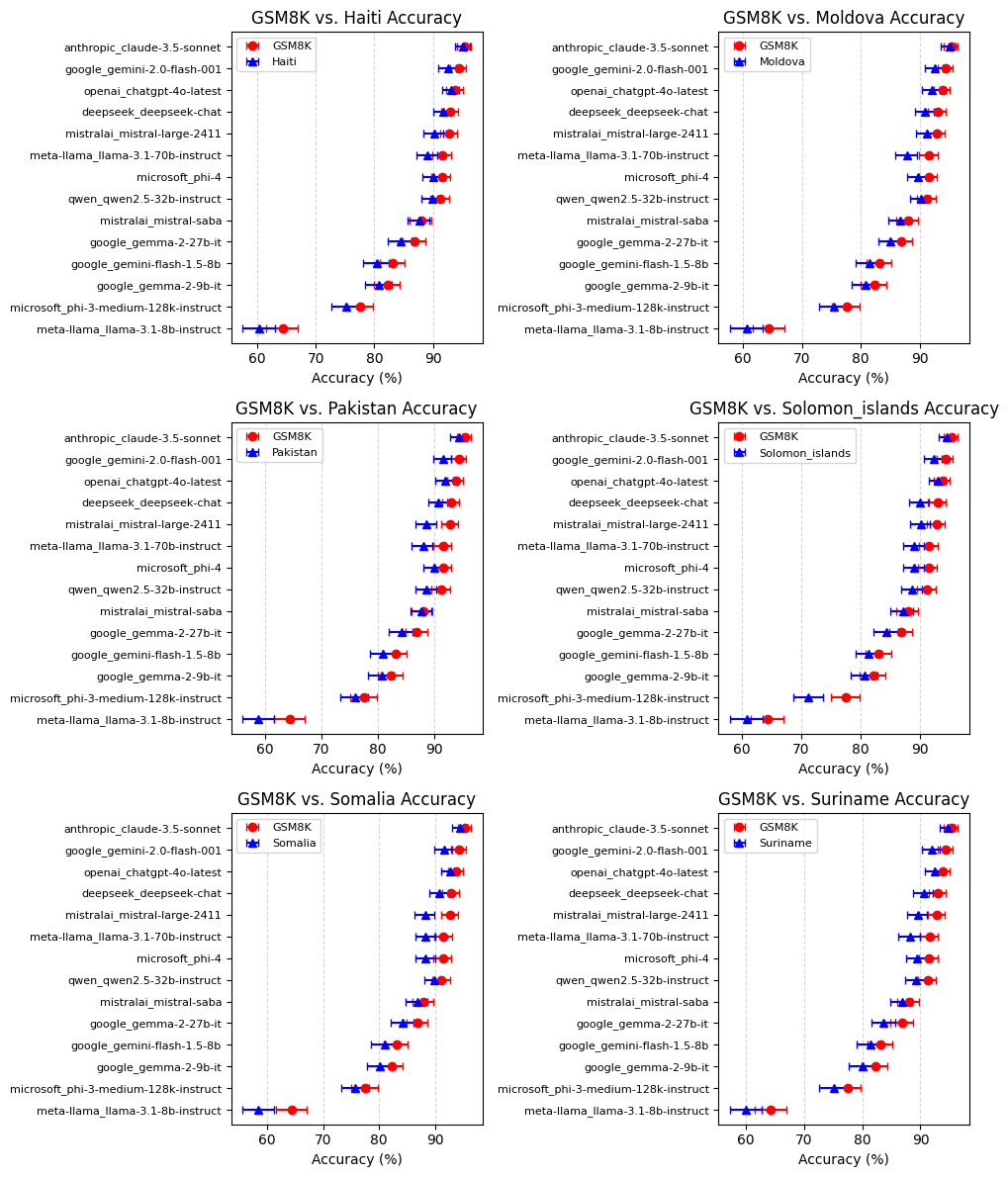}
    \caption{Accuracy Comparison of GSM8K vs culturally variant versions of GSM8K across various models }
    \label{fig:3}
\end{figure}

To analyze the impact of cultural modifications on the mathematical reasoning ability of LLMs, we compare their strict accuracy on the original GSM8K dataset and its six culturally adapted variants for each model. The results are presented in Figure~\ref{fig:3}, where each subplot compares the performance of the model in GSM8K (red) with its corresponding cultural dataset (blue). Each point represents strict accuracy, calculated under a consistency criterion where a model is only marked correct if all three outputs match the ground truth. These points in all the plots are accompanied by horizontal lines indicating the 95\% confidence interval, computed using the Wilson score method. These intervals reflect the uncertainty around the strict accuracy estimates and help visualize whether observed differences between GSM8K and its cultural variants are likely statistically meaningful. Across all models and datasets, we observe a consistent pattern: models perform better on GSM8K than on the culturally modified versions. This is visually evident by the leftward shift of blue dots compared to red ones, indicating a drop in accuracy when problems are reframed in different cultural contexts. The drop is relatively smaller for Haiti and Moldova, while more pronounced for Pakistan, Somalia, Solomon Islands, and Suriname. These results suggest that even when the underlying mathematical logic remains unchanged, cultural framing can significantly influence model performance.

This variation suggests that certain cultural shifts disrupt reasoning more than others, possibly due to differences in entity recognition, familiarity with cultural references, or exposure to training data.

Claude 3.5 Sonnet, GPT-4o, Gemini 2.0, and Qwen 2.5-32B showed the smallest accuracy drop, implying better generalization across cultural contexts. These models retained strong reasoning abilities despite cultural variations. While, Meta LLaMA 3.1-8B, Microsoft Phi-3 Medium, Gemma-2-9B showed substantial accuracy reductions, indicating difficulty adapting to cultural modifications. Phi-3 Medium and LLaMA 3.1-8B exhibited significant drops on Pakistan, Solomon Islands, and Somalia datasets, suggesting that these specific variations posed greater challenges for smaller models.

Solomon Islands dataset resulted in the largest performance gap across models. Meta LLaMA 3.1-8B and Microsoft Phi-3 Medium exhibited the most drastic accuracy reductions on Pakistan and Somalia, where reasoning failures were more prominent. Mistral models performed relatively well across all datasets, particularly Mistral Saba, which is trained on a highly diverse dataset encompassing linguistic and cultural nuances from the Middle East and South Asia, but still exhibited noticeable drops, particularly on Suriname and Pakistan.

For detailed numerical values corresponding to these accuracy scores, including confidence intervals, refer to Table~\ref{tab:3} in (Appendix B).

\clearpage

\subsubsection{Performance Gap Analysis}
\begin{figure}[H]
    \centering
    \includegraphics[width=1\textwidth]{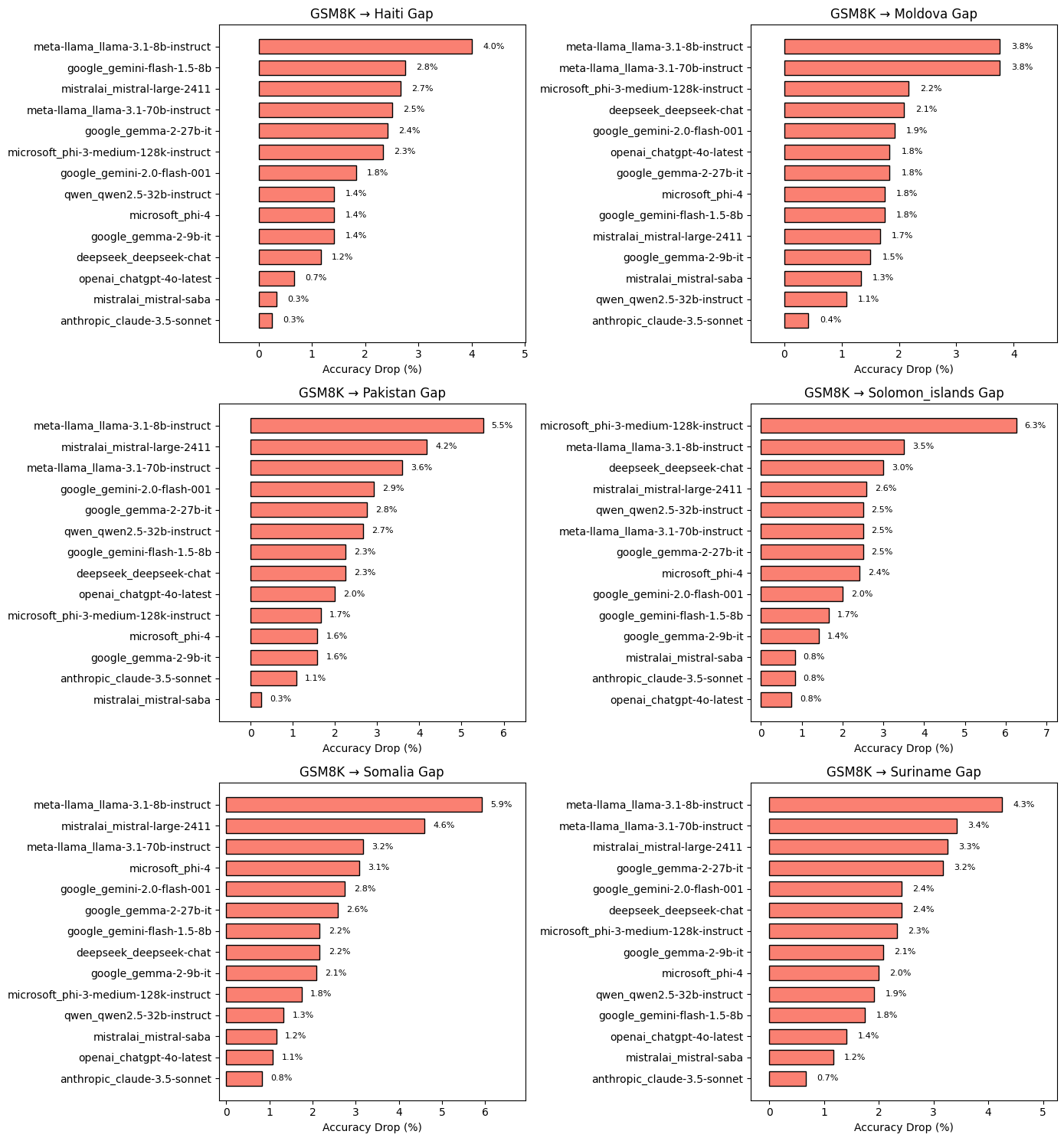}
    \caption{Performance Gap of Models across various culturally adapted GSM8K variants}
    \label{fig:4}
\end{figure}

We compute a performance gap between the GSM8K test set and its culturally adapted versions. The performance gap is calculated by subtracting the accuracy of each model in a cultural variant from its accuracy in the original GSM8K. A higher value indicates a greater drop in performance when faced with culturally adapted math problems and vice versa. This gap analysis allows us to identify which models are more sensitive to cultural variations and which ones generalize better across diverse linguistic or contextual settings. The bar charts in Figure~\ref{fig:4} display this accuracy drop across multiple cultural variants for each model, with the models listed on the y-axis and the magnitude of the accuracy drop shown on the x-axis. Across all graphs, Meta-LLaMA 3.1-8B-Instruct consistently shows the highest accuracy drop for multiple datasets. This indicates that this model struggles the most with math reasoning in diverse cultural contexts. Anthropic Claude 3.5-Sonnet consistently shows the smallest accuracy drop, meaning it maintains the most stable performance across datasets.

In Haiti for instance, Meta-LLaMA 3.1-8B-Instruct has the highest accuracy drop (4.0\%), showing that it struggles with numerical reasoning in Haitian contexts. Most models have a drop of 1-3\%, with Claude 3.5-Sonnet dropping only 0.3\%. Meta-LLaMA 3.1-8B and 70B have the highest accuracy drop on Moldova(~3.8\%), followed by Microsoft Phi-3 Medium. Claude 3.5-Sonnet remains the most stable, with only a 0.4\% drop. This shows that models that are heavily optimized for English-centric training data tend to show accuracy drops on Moldova, probably due to insufficient exposure of cultural reasoning patterns in training data. While models such as Claude 3.5-Sonnet demonstrate stronger generalization across cultural variants, the specific factors driving this robustness, whether model scale, training data diversity, or stronger mathematical reasoning capabilities, cannot be extracted from our experiments. Detailed values can be found in Table~\ref{tab:4} and Table~\ref{tab:5} (Appendix B).  

Claude 3.5 consistently outperforms other models across various datasets and is notably robust in mathematical reasoning. It achieved a score of 71.1\% on the MATH benchmark and an impressive 91.6\% on multilingual math tasks (MGSM) \citep{anthropic2024claude35}. However, when cultural and linguistic variations are introduced, the performance dynamics changes slightly.  Interestingly, Mistral Saba, despite not having a dedicated mathematical reasoning benchmark and not being explicitly tuned for math, handles the Pakistan variant and some other cultural adaptations better. This may be attributed to its training on data from Middle Eastern and South Asian sources. This indicates that even if a model isn't specifically designed for mathematical reasoning, having a deeper understanding of a region's cultural context and reasoning patterns can help it perform better on math tasks within that specific cultural framework.

There could be  other possible reasons for the accuracy drop of models when the context is shifted. Different languages and cultural terminologies may lead to variations in tokenization. If the model's tokenizer is not well-suited to the culturally adapted language, a single concept might be represented by more tokens, potentially increasing input complexity. The model may also have fewer training examples with the specific terms used in the adapted problems \citep{petrov2023language}, \citep{dang2024tokenization}. This suggests that models are not entirely language-agnostic for mathematical reasoning performance, even if they are multilingual. Their mathematical reasoning performance may correlate with how efficiently their tokenizers can represent different languages and cultural vocabularies, though establishing a direct causal link falls outside the scope of this black-box evaluation. As shown in Figure~\ref{fig:9} (Appendix B), adapting names from ‘Amalia, Megan, and Dior’ to ‘Aleskandra, Nicolae, and Albert’ changes the total number of tokens and characters. In the original text, we have 104 tokens and 452 characters, while the Moldovan-adapted version yields 109 tokens and 477 characters. This variation reflects how the tokenizer handles different linguistic structures and represents one potential contributing factor among several possible explanations for the observed performance differences.

Similarly, mathematical reasoning isn't always culturally neutral. The way a problem is framed, even with the same numbers, can subtly influence how someone may  approach it. Different cultures might have different common-sense assumptions or preferred methods for organizing information \citep{meng2022effects}, \citep{tajika2004differences}. This implies that LLMs aren't just learning to solve math problems; they're also learning to solve them in a particular way, based on the dominant problem-solving styles in their training data. Cultural versions might require slightly different reasoning pathways that the model is less familiar with. If the pre-training data is heavily biased towards certain cultures or problem-solving styles, the model will naturally perform better in those familiar contexts. This underscores the importance of diverse and representative pre-training datasets.

\subsubsection{Statistical Significance Testing}
To statistically assess whether LLMs perform differently on culturally adapted math questions compared to the original GSM8K test set, we conduct McNemar tests using one-to-one aligned question pairs. The results reveal a clear pattern: models like LLaMA 3.1-70B, Gemini Flash 2.0, and Mistral Large 2411 consistently exhibit statistically significant performance drops (p $<$ 0.01 across most datasets), with high b-values indicating that these models frequently answered questions correctly on GSM8K but failed on their culturally adapted versions. This supports the rejection of the null hypothesis, suggesting these models struggle more in mathematical reasoning when cultural contexts shift.

In contrast, Claude 3.5 and Mistral Saba generally show no significant performance difference (p $>$ 0.05 in most cases), with balanced b/c values, indicating that their performance is more stable across cultural contexts, and we fail to reject the null hypothesis for these models.

A detailed breakdown of McNemar test statistics, p-values, and b/c counts for each model is provided in Table~\ref{tab:6} (Appendix B).
\subsubsection{Qualitative Error Analysis}
\label{4.2.4}
We conduct a detailed qualitative error analysis to better understand how LLMs handle culturally adapted math word problems. The qualitative error analysis was conducted through systematic manual inspection of model outputs across all 14 models and all six cultural variants. Full methodology and examples are provided in Appendix B Section \ref{appendix:qualitative-error-analysis}.  Our analysis reveals three major patterns of reasoning failure across models:
\begin{itemize}
    \item Models often struggle with numerical reasoning when using less familiar currency units (e.g., Haitian Gourde, HTG). For instance, some models treat 0.1 HTG as though it were 1 HTG, whereas they correctly handle 0.1 USD, likely because HTG is rarely used in decimals due to inflation and rounding practices, causing the models to miss consistent arithmetic across currency formats.
    \item Replacing “wife” with unfamiliar family terms like Jija (Pakistani) or Tambu man (Solomon Islands) often led models to miscalculate. While they correctly process “husband-wife,” they struggle with non-Western family structures, pointing to errors in how they interpret different cultural entities.
    \item Sometimes the models fail to link culturally specific terms to their real meanings. For example, substituting local animal names in a counting problem lead the models to default to incorrect assumptions, suggesting when confronted with unfamiliar cultural words, they may be relying on learned patterns from their original training data.
\end{itemize}
 These insights highlight how cultural context can introduce variability in reasoning even when the underlying math remains unchanged. Examples and detail of these errors are provided in Figures~\ref{fig:10}, \ref{fig:11}, and \ref{fig:12} (Appendix B)

\subsubsection{Quantitative Error Analysis}
While Section ~\ref{4.2.4} illustrates specific failure patterns through representative examples, we systematically quantify the prevalence of different error types across all models and cultural variants. We analyze 18,887 error instances identified using the strict consistency criterion described in Section 
 ~\ref{4.1}, where at least one of the three model runs produced an incorrect answer. These errors are aggregated across all 14 evaluated models and all seven datasets, the original GSM8K and its six cultural variants, and include all incorrect responses regardless of whether the model answered the original GSM8K question correctly or not. Each error is categorized into six types, as shown in the Table ~\ref{tab:error_distribution}, using an LLM-assisted annotation process, which is validated through manual review of 50 randomly sampled cases (detailed methodology in Appendix B Section ~\ref{quantitative-error-analysis}).

\begin{table}[h]
\centering
\begin{tabular}{|l|c|c|}
\hline
\textbf{Error Category} & \textbf{Count} & \textbf{Percentage (\%)} \\
\hline
Mathematical Reasoning  error & 10,331 & 54.70 \\
\hline
Calculation error & 6,523 & 34.54 \\
\hline
Relationship misunderstanding & 851 & 4.51 \\
\hline
Unit / currency error & 746 & 3.95 \\
\hline
Problem misinterpretation & 292 & 1.55 \\
\hline
Other / unclear & 144 & 0.76 \\
\hline
\textbf{Total} & \textbf{18,887} & \textbf{100.00} \\
\hline
\end{tabular}
\caption{Overall distribution of error types across all models and cultural variants.}
\label{tab:error_distribution}
\end{table}

Table ~\ref{tab:error_distribution} presents the overall distribution of error types. The analysis reveals that mathematical reasoning errors constitute the majority of failures (54.7\%), followed by calculation errors (34.5\%). Notably, culturally-specific errors, relationship misunderstandings (4.5\%) and unit/currency errors (4.0\%), comprise a relatively small proportion of total errors. This distribution quantifies the error patterns identified in Section ~\ref{4.2.4}, revealing that while explicitly culturally-specific errors (relationship and currency) represent 8.5\% of failures, the broader impact of cultural adaptation on model reasoning is reflected across all error categories. Representative examples of each error type are provided in Appendix B Section \ref{appendix:qualitative-error-analysis}.
A complementary analysis examining which cultural entity categories are disproportionately associated with incorrect responses is provided in Appendix B Section \ref{Error Distribution by Entity Category}. 
 
\section{Conclusion}
In this work, we investigate whether LLMs can maintain stable performance in solving mathematical problems in diverse cultural contexts. To do so, we adapted the GSM8K dataset by incorporating cultural elements, thus creating six regional variants alongside the original dataset. Our evaluation of these variants reveals that while the mathematical principles underlying GSM8K remain unchanged, the introduction of cultural variations does affect model performance.

The results show that larger models, such as Anthropic Claude 3.5-Sonnet and GPT-4o, tend to generalize better across these diverse contexts; however, they still experience noticeable performance drops when faced with culturally adapted math problems. In contrast, models like Mistral Saba, despite not being explicitly tuned for math, demonstrate improved performance in regions where they have been exposed to local data, suggesting that cultural familiarity can enhance mathematical problem solving. Furthermore, our analysis indicates that changes in cultural context may also pose challenges to the tokenisation process, as variations in vocabulary and linguistic structures across different regions can lead to differences in how input text is tokenised.

In general, our study underscores the importance of considering cultural context when evaluating the mathematical reasoning capabilities of LLMs. Beyond evaluation, the culturally adapted benchmarks introduced in this study could serve as a foundation for future work in training more culturally robust LLMs, for instance, through targeted data augmentation strategies that expose models to underrepresented cultural contexts during fine-tuning. The dataset creation pipeline presented here is also replicable and extensible, offering a template for developing similar benchmarks across additional cultures and languages.



\appendix
\setcounter{figure}{0}
\renewcommand{\thefigure}{A\arabic{figure}}
\setcounter{table}{0}
\renewcommand{\thetable}{A\arabic{table}}
\renewcommand{\thesubsection}{\arabic{subsection}}

\section*{Appendix A: Dataset Creation}
\addcontentsline{toc}{section}{Appendix A: Dictionary Creation}
\label{appendix:a}

\subsection{Prompt for Cultural Entities Recognition}
\label{appendix:prompt-cultural-entities-recognition}
\begin{figure}[H]
    \centering
    \includegraphics[width=0.8\textwidth]{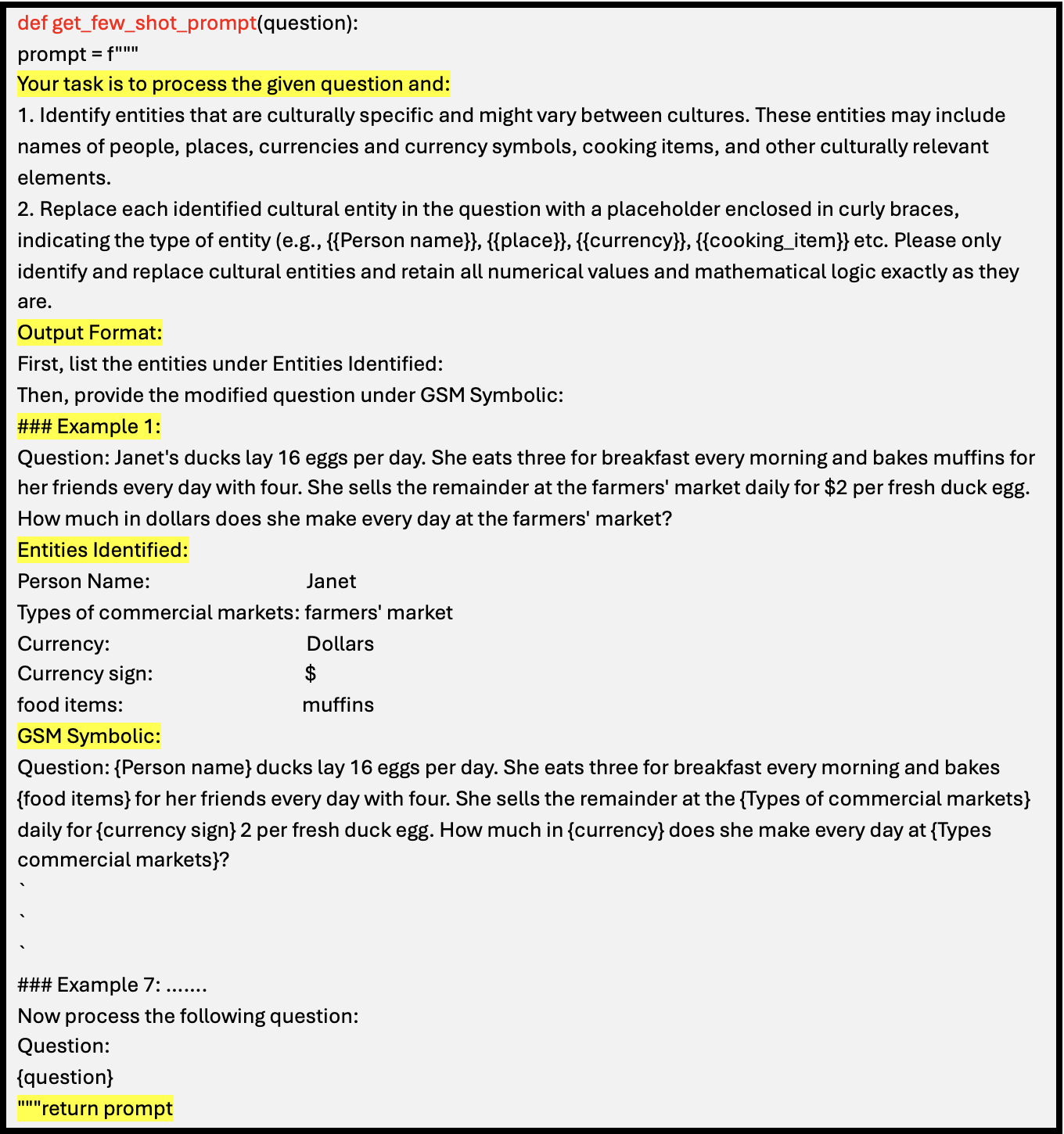}
    \caption{Prompt for Cultural Entities Recognition}
    \label{fig:5}
\end{figure}

\clearpage
\begin{figure}[H]
    \centering
    \includegraphics[width=0.8\textwidth]{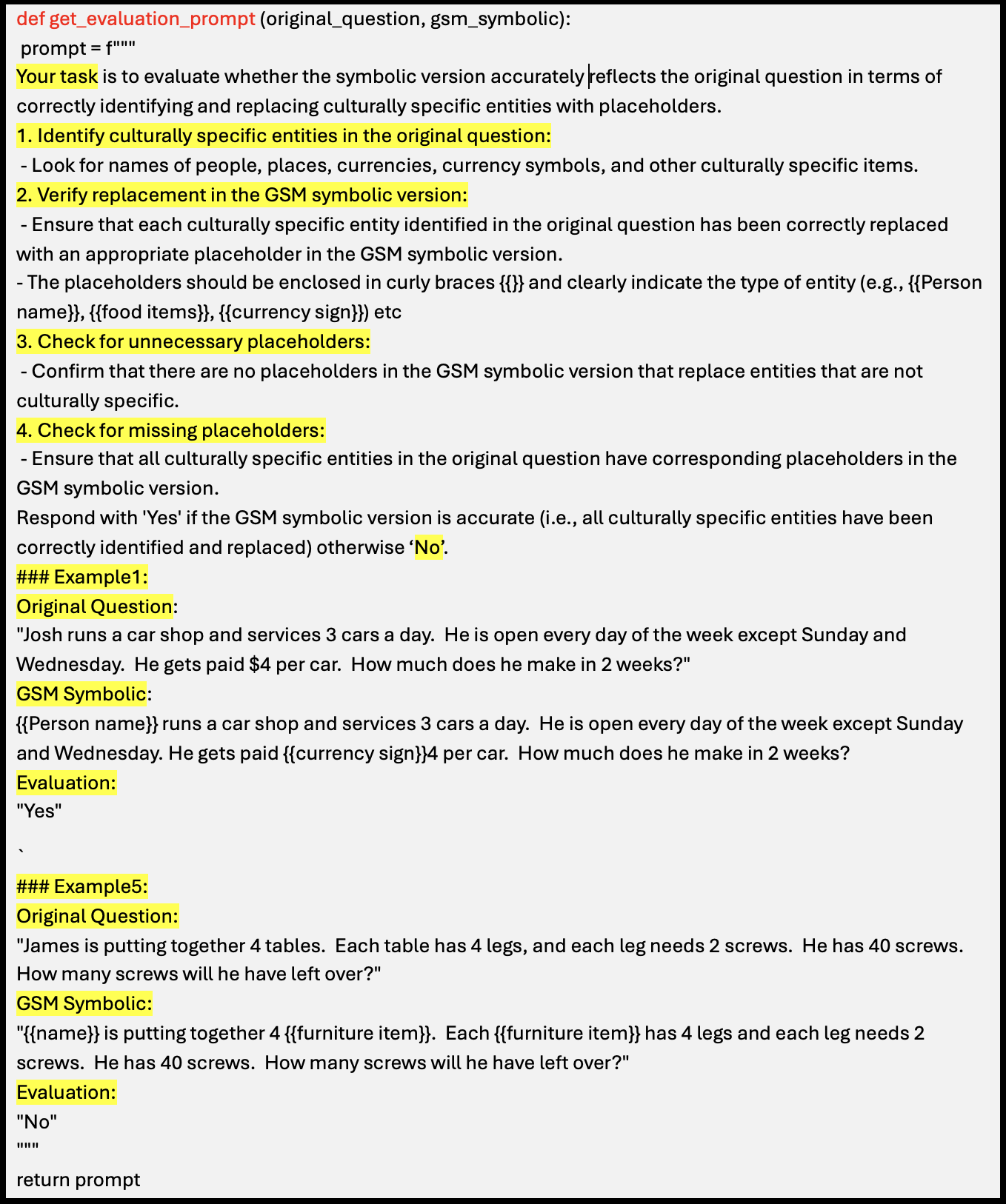}
    \caption{Prompt for Recognized Cultural Entities Evaluation}
    \label{fig:6}
\end{figure}

\subsection{Validation of Cultural Entity Recognition Quality}
\label{Validation of Cultural Entity Recognition Quality}
\subsubsection{Sampling Methodology}
We randomly sampled 130 questions from the 1,198 culturally adaptable questions in our dataset, representing approximately 11\% of the total. This sample size provides sufficient statistical power for quality assessment while remaining feasible for detailed manual review. The random sampling was performed without stratification to ensure an unbiased representation of the dataset's diversity in terms of question complexity, 
number of entities per question, and entity types.

\subsubsection{Annotation Guidelines and Evaluation Criteria}
Two independent human annotators (authors of this paper with expertise in the dataset creation process) were tasked with evaluating GPT-4o's entity recognition output. For each of the 130 sampled questions, annotators were instructed to answer a binary question: ``Did GPT-4o correctly identify all cultural entities in this question without omissions or errors?"
The evaluation criteria were defined as follows:
\begin{itemize}
    \item \textbf{Yes (Correct):} GPT-4o identified all cultural entities present in the question with no omissions and no incorrect identifications.
    \item \textbf{No (Incorrect):} GPT-4o either (1) missed at least one cultural entity, (2) incorrectly identified a non-entity as a cultural entity, or (3) misclassified the entity type.
\end{itemize}

This strict, question-level evaluation approach ensures that only questions with perfect entity recognition are marked as correct, providing a conservative estimate of GPT-4o's performance.

\subsubsection{Validation Results}
The two human annotators as well as GPT-4o, evaluated all 130 sampled questions independently. The results are summarized in Table ~\ref{tab:validation_results}:

\begin{table}[h]
\centering
\begin{tabular}{|l|c|c|c|}
\hline
\textbf{Evaluator} & \textbf{Correct (Yes)} & \textbf{Incorrect (No)} & \textbf{Accuracy (\%)} \\
\hline
Human Annotator 1 & 96 & 34 & 73.8 \\
\hline
Human Annotator 2 & 103 & 27 & 79.2 \\
\hline
GPT-4o (Self-evaluation) & 86 & 44 & 66.2 \\
\hline
\end{tabular}
\caption{Validation results for cultural entity recognition on 130 sampled questions.}
\label{tab:validation_results}
\end{table}

The validation results show that the two human annotators achieved question-level accuracy of 73.8\% and 79.2\%, respectively, indicating that GPT-4o successfully identified all cultural entities without errors in approximately three out of four questions. Additionally, we employed GPT-4o itself as a third evaluator to assess its own output, which yielded 86 correct identifications (66.2\% accuracy), suggesting it applied stricter criteria when assessing its own output. 

\subsubsection{Correction and Iterative Refinement}
All questions identified as incorrect by any of the evaluators were manually reviewed to identify the specific errors in GPT-4o's entity recognition. Errors fell into two categories: placeholder naming inconsistencies, addressed through the prompt refinement process described in Appendix A Section ~\ref{appendix:entity-correction}, and broader entity recognition errors including missed entities and incorrect entity type classifications, addressed through the multi-layer correction process described below.

To address broader entity recognition errors across the full dataset, we employed GPT-4o as an automated evaluator on all 1,198 culturally adaptable questions. Unlike its role as a third evaluator in the 130-question validation study, here GPT-4o was explicitly instructed with the specific error patterns identified during human validation, including missed entities and incorrect entity type classifications, and applied systematically across every question in the dataset.

In parallel, we refined the entity recognition 7-shot prompt iteratively based on the error patterns discovered during both the human validation and GPT-4o automated evaluation. Each refined version of the prompt was reapplied to the complete dataset, with subsequent evaluation cycles used to verify improvement. This iterative prompt refinement addressed both placeholder naming inconsistencies, detailed in Appendix A Section ~\ref{appendix:entity-correction}, and broader entity recognition errors across the full dataset.

Finally, manual human inspection was conducted across the full dataset by the authors, drawing on our cultural familiarity with the target regions, particularly Pakistan, to identify and correct culturally sensitive errors that automated evaluation could not reliably assess. This was especially important for entity types requiring cultural understanding, such as regional family terms, local food items, and culturally specific place names. The base evaluation prompt structure used to guide GPT-4o as a judge throughout this process is provided in Appendix A Figure ~\ref{fig:6}, with targeted refinements applied iteratively based on errors discovered in each correction cycle.

The validation study confirms that while GPT-4o's initial entity recognition achieved reasonable accuracy (73-79\% as judged by human annotators), manual oversight and iterative correction, the multi-layer correction process, combining automated evaluation, iterative prompt refinement, and manual human inspection, was essential to ensure the quality and reliability of our culturally adapted datasets.

\subsection{Iterative Correction of Placeholder Naming Inconsistencies}
\label{appendix:entity-correction}
During the validation study described in Appendix A Section~\ref{Validation of Cultural Entity Recognition Quality}, we employed a 5-shot \textbf{evaluation} prompt (Figure ~\ref{fig:6}) to evaluate GPT-4o's entity recognition output. This evaluation process revealed a naming inconsistency issue in GPT-4o's placeholder assignments. While GPT-4o correctly identified cultural entities, it used inconsistent placeholder names for the same entity type across different questions. For example, person names were variously labeled as \textit{name}, \textit{common name}, or \textit{Person name}. Similarly, food-related entities appeared as \textit{food item}, \textit{common food items}, \textit{types of food}, and \textit{cooking items}. 

This inconsistency posed a problem for dictionary creation, as different placeholder names for the same entity type would create redundant keys with identical value sets. To address this issue, we modified the 7-shot prompt (Figure~\ref{fig:5}) to enforce consistent placeholder naming conventions and conducted additional iterations of entity recognition.

\subsection{Recognized Cultural Entities by GPT-4o}
\label{appendix:recognized-cultural-entities}

\begin{table}[H]
\centering
\small
\renewcommand{\arraystretch}{1.2}

\begin{minipage}[t]{0.48\textwidth}
\vspace{0pt}
\begin{tabular}{>{\raggedright\arraybackslash}p{2.5cm}>{\raggedright\arraybackslash}p{2.9cm}>{\raggedright\arraybackslash}p{0.8cm}}
\hline\hline
\textbf{Category} & \textbf{Entity Type} & \textbf{\%} \\
\hline\hline

People and Roles
& Person name & 87.1\% \\
& Family member & 4.9\% \\
& Types of common jobs & 2.0\% \\
& Profession & 1.0\% \\
\hline

Places and Locations
& Types of commercial establishments & 5.7\% \\
& City name & 1.8\% \\
& Types of places & 1.0\% \\
& Restaurant name & 0.8\% \\
& School name & 0.7\% \\
& Types of entertainment places & 0.5\% \\
& Types of houses & 0.4\% \\
& Common places & 0.2\% \\
& Religious place & 0.2\% \\
& Village names & 0.1\% \\
& Recreation places & 0.1\% \\
& Cultural landmark & 0.1\% \\
& Government body & 0.1\% \\
\hline

Food and Drink
& Food items & 11.2\% \\
& Cooking item & 2.1\% \\
& Types of beverages & 1.2\% \\
& Types of pastries/local desserts & 0.1\% \\
& Types of tea & 0.1\% \\
\hline

Education
& School subject & 0.4\% \\
& Types of classes & 0.3\% \\
& Types of teacher & 0.3\% \\
\hline

Clothing and Appearance
& Clothing items & 1.5\% \\
& Common clothing items & 1.5\% \\
& Types of flowers & 0.4\% \\
& Types of scents & 0.1\% \\
\hline\hline
\end{tabular}
\end{minipage}
\hfill
\begin{minipage}[t]{0.48\textwidth}
\vspace{0pt}
\begin{tabular}{>{\raggedright\arraybackslash}p{2.5cm}>{\raggedright\arraybackslash}p{2.9cm}>{\raggedright\arraybackslash}p{0.8cm}}
\hline\hline
\textbf{Category} & \textbf{Entity Type} & \textbf{\%} \\
\hline\hline

Culture and Arts
& Types of events & 2.2\% \\
& Cultural event & 1.2\% \\
& Types of family events & 1.0\% \\
& Holiday & 0.5\% \\
& Types of books & 0.3\% \\
& Mythical character & 0.3\% \\
& Types of shows & 0.2\% \\
& Types of musical compositions & 0.2\% \\
& Recreation activity & 0.2\% \\
& Types of dance & 0.1\% \\
& Cultural dance style & 0.1\% \\
& Cultural songs & 0.1\% \\
& Types of games & 0.1\% \\
\hline

Commerce and Economy
& Currency sign & 31.8\% \\
& Currency & 20.7\% \\
& Company names & 0.5\% \\
& Common brand name & 0.4\% \\
& Types of goods merchant purchase & 0.1\% \\
& Online shopping platforms & 0.1\% \\
& Appliances & 0.1\% \\
\hline

Other
& Animal & 2.3\% \\
& Common type of sport & 1.3\% \\
& Types of vehicles & 0.2\% \\
& Language & 0.1\% \\
& Newspaper names & 0.1\% \\
\hline\hline
\end{tabular}
\end{minipage}

\caption{Cultural Entity Types by Category with Occurrence Percentages}
\label{tab:2}
\smallskip
\begin{minipage}{\textwidth}
\small Table ~\ref{tab:2} presents the 54 cultural entity types identified across the 1,198 culturally adaptable questions, organised into eight broad categories. The percentage indicates the proportion of questions in which each entity type appears at least once. Person name is the most frequently occurring entity type, present in 87.1\% of questions, followed by currency sign (31.8\%) and currency (20.7\%), reflecting the nature of GSM8K as a math word problem dataset involving people and monetary values. Most other entity types appear in fewer than 5\% of questions.
\end{minipage}
\end{table}

\subsection{Dictionary}
\label{appendix:dictionary}

\begin{figure}[H]
    \centering
    \includegraphics[width=0.6\textwidth]{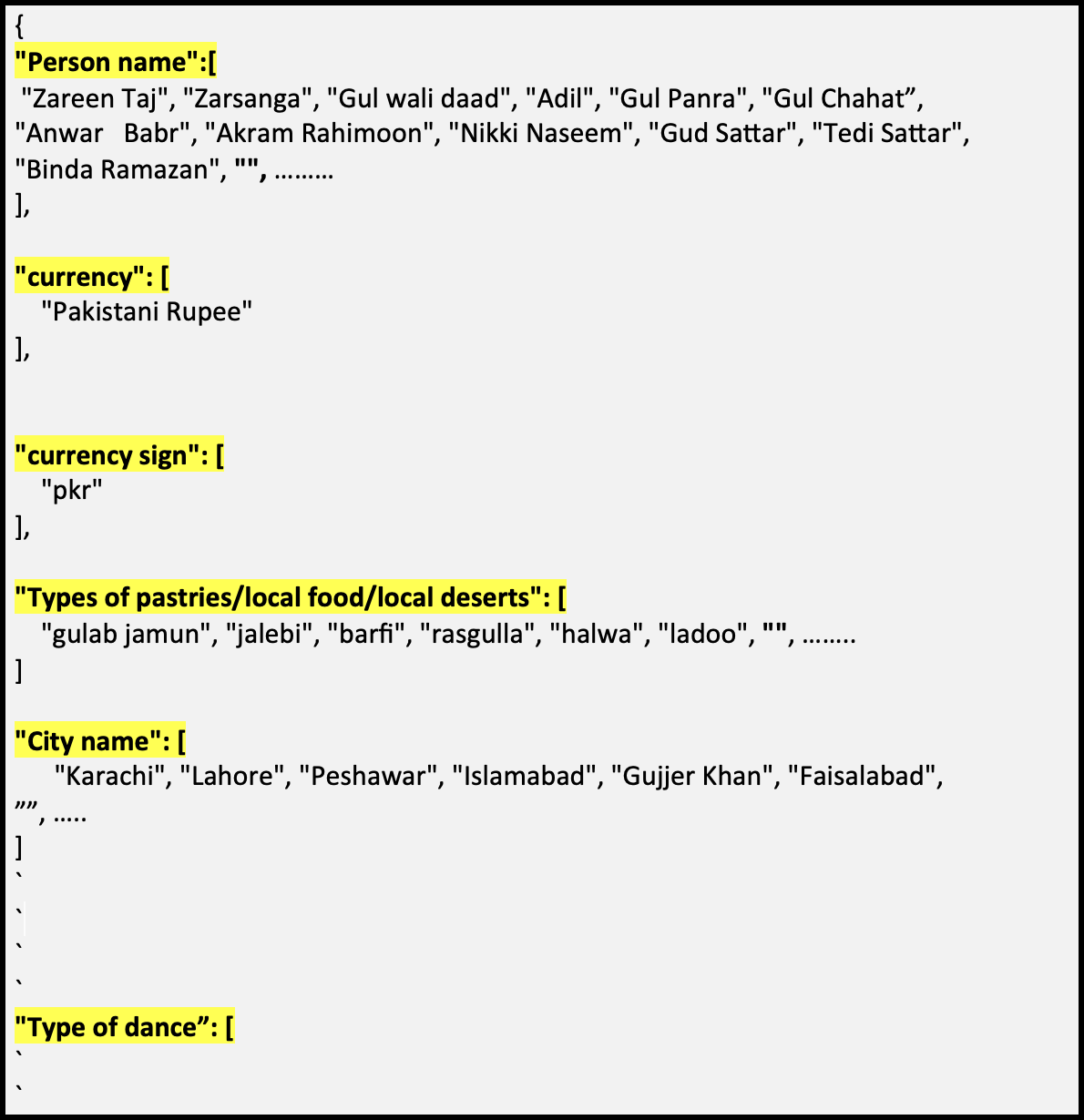}
    \caption{Screenshot of a Dictionary}
    \label{fig:7}
\end{figure}

\section*{Appendix B: LLMs Evaluation}
\addcontentsline{toc}{section}{Appendix B: LLM Evaluation}
\label{appendix:b}

\subsection{Prompt}
\label{appendix:prompt}

\begin{figure}[H]
    \centering
    \includegraphics[width=0.7\textwidth]{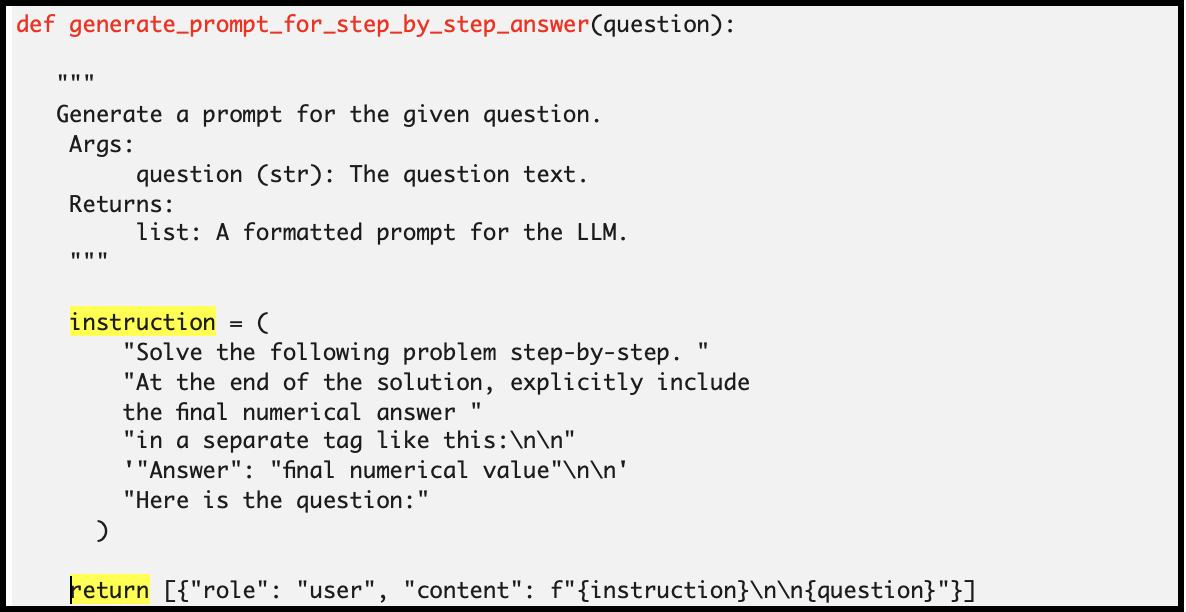}
    \caption{Prompt for LLMs Evaluation on all Datasets}
    \label{fig:8}
\end{figure}

Our prompt is straightforward, it does not instruct the LLM to solve math problems in any specific way, nor does it emphasize math, numerical reasoning, or cultural elements. Instead, it simply guides the LLM to solve the problem step by step. The actual math problem is then introduced within this structured approach. We also use the same prompt on all of  the models for all the datasets. 

By following this simple prompting method, we ensure that no external factors influence the LLM’s response. The model focuses solely on solving the given mathematical problem, independent of any cultural or contextual elements. This allows us to obtain clean responses from the models when we make cultural modifications to the GSM8K questions.

\subsection{Models Accuracy Across Cultural Variants}
\label{appendix:models-accuracy-across-cultural-variants}

\begin{table}[h]
    \centering
    \footnotesize 
    \renewcommand{\arraystretch}{0.9} 
    \setlength{\tabcolsep}{3pt} 
    \begin{tabular}{lccccccc}
        \toprule
        \textbf{Model} & \textbf{G8K} & \textbf{Hti} & \textbf{Mld} & \textbf{Pak} & \textbf{Sol} & \textbf{Som} & \textbf{Sur} \\
        \midrule
        C3.5  & 0.95 & 0.95 & 0.94 & 0.94 & 0.94 & 0.94 & 0.94 \\
              & (0.94-0.96) & (0.93-0.96) & (0.93-0.96) & (0.92-0.95) & (0.93-0.95) & (0.93-0.95) & (0.93-0.95) \\
        \midrule
        DSeek  & 0.92 & 0.91 & 0.90 & 0.90 & 0.89 & 0.90 & 0.90 \\
               & (0.91-0.94) & (0.90-0.93) & (0.89-0.92) & (0.88-0.92) & (0.88-0.91) & (0.89-0.92) & (0.88-0.92) \\
        \midrule
        G2.0  & 0.94 & 0.92 & 0.92 & 0.91 & 0.92 & 0.91 & 0.91 \\
              & (0.92-0.95) & (0.90-0.93) & (0.90-0.93) & (0.89-0.92) & (0.90-0.93) & (0.89-0.93) & (0.90-0.93) \\
        \midrule
        G1.5  & 0.83 & 0.80 & 0.81 & 0.80 & 0.81 & 0.80 & 0.81 \\
              & (0.80-0.85) & (0.78-0.82) & (0.79-0.83) & (0.78-0.83) & (0.79-0.83) & (0.78-0.83) & (0.79-0.83) \\
        \midrule
        G27B  & 0.86 & 0.84 & 0.85 & 0.84 & 0.84 & 0.84 & 0.83 \\
              & (0.84-0.88) & (0.82-0.86) & (0.82-0.86) & (0.81-0.86) & (0.82-0.86) & (0.82-0.86) & (0.81-0.85) \\
        \midrule
        G9B  & 0.82 & 0.80 & 0.80 & 0.80 & 0.80 & 0.80 & 0.80 \\
             & (0.79-0.84) & (0.78-0.82) & (0.78-0.82) & (0.78-0.82) & (0.78-0.82) & (0.77-0.82) & (0.77-0.82) \\
        \midrule
        L70B  & 0.91 & 0.89 & 0.87 & 0.87 & 0.89 & 0.88 & 0.88 \\
             & (0.89-0.93) & (0.87-0.90) & (0.85-0.89) & (0.86-0.89) & (0.87-0.90) & (0.86-0.90) & (0.86-0.89) \\
        \midrule
        L8B  & 0.64 & 0.60 & 0.60 & 0.58 & 0.60 & 0.58 & 0.60 \\
            & (0.61-0.67) & (0.57-0.63) & (0.57-0.63) & (0.56-0.61) & (0.58-0.63) & (0.55-0.61) & (0.57-0.62) \\
        \midrule
        P3M  & 0.77 & 0.75 & 0.75 & 0.75 & 0.71 & 0.75 & 0.75 \\
            & (0.75-0.79) & (0.72-0.77) & (0.72-0.77) & (0.73-0.78) & (0.68-0.73) & (0.73-0.78) & (0.72-0.77) \\
        \midrule
        P4  & 0.91 & 0.90 & 0.89 & 0.89 & 0.89 & 0.88 & 0.89 \\
           & (0.89-0.92) & (0.88-0.91) & (0.87-0.91) & (0.88-0.91) & (0.87-0.90) & (0.86-0.90) & (0.87-0.91) \\
        \midrule
        M2411  & 0.92 & 0.90 & 0.91 & 0.88 & 0.90 & 0.88 & 0.89 \\
               & (0.91-0.94) & (0.88-0.91) & (0.89-0.92) & (0.86-0.90) & (0.88-0.91) & (0.86-0.89) & (0.87-0.91) \\
        \midrule
        MSaba  & 0.87 & 0.87 & 0.86 & 0.87 & 0.87 & 0.86 & 0.86 \\
               & (0.86-0.89) & (0.85-0.89) & (0.84-0.88) & (0.85-0.89) & (0.85-0.88) & (0.84-0.88) & (0.84-0.88) \\
        \midrule
        G4o  & 0.93 & 0.93 & 0.91 & 0.91 & 0.93 & 0.92 & 0.92 \\
            & (0.92-0.95) & (0.91-0.94) & (0.90-0.93) & (0.90-0.93) & (0.91-0.94) & (0.91-0.94) & (0.90-0.93) \\
        \midrule
        Q32B  & 0.91 & 0.89 & 0.90 & 0.88 & 0.88 & 0.89 & 0.89 \\
             & (0.89-0.92) & (0.87-0.91) & (0.88-0.91) & (0.86-0.90) & (0.86-0.90) & (0.88-0.91) & (0.87-0.90) \\
        \bottomrule
    \end{tabular}
    \caption{Accuracy Scores Across Models and Datasets. Values in parentheses indicate confidence intervals (CI). C3.5 = anthropic\_claude-3.5-sonnet, DSeek = deepseek\_deepseek-v3, G2.0 = google\_gemini-2.0-flash-001, G1.5 = google\_gemini-flash-1.5-8b, G27B = google\_gemma-2-27b-it, G9B = google\_gemma-2-9b-it, L70B = meta-llama\_llama-3.1-70b-instruct, L8B = meta-llama\_llama-3.1-8b-instruct, P3M = microsoft\_phi-3-medium-128k-instruct, P4 = microsoft\_phi-4, M2411 = mistralai\_mistral-large-2411, MSaba = Mistral Saba, G4o = chatgpt-4o-latest, Q32B = qwen2.5-32b-instruct. G8K = GSM8K, Hti = HaiGSM8K, Mld = MolGSM8K, Pak = PakGSM8K, Sol = SolIGSM8K, Som = SomGSM8K, Sur = SurGSM8K.}
    \label{tab:3}
\end{table}

\subsection{Performance Gap}
\label{appendix:performance-gap}

\begin{table}[H]
    \centering
    \renewcommand{\arraystretch}{1.2} 
    \setlength{\tabcolsep}{4pt} 
    \scriptsize 
    \begin{tabular}{lcccccc}
        \toprule
        \textbf{Model} & \textbf{Hti Gap} & \textbf{Mld Gap} & \textbf{Pak Gap} & \textbf{Sol Gap} & \textbf{Som Gap} & \textbf{Sur Gap} \\
        \midrule
        Claude 3.5 & 0.0025 & 0.0042 & 0.0109 & 0.0083 & 0.0083 & 0.0067 \\
        DeepSeek & 0.0117 & 0.0209 & 0.0225 & 0.0301 & 0.0217 & 0.0242 \\
        Gemini 2.0 & 0.0184 & 0.0192 & 0.0292 & 0.0200 & 0.0275 & 0.0242 \\
        Gemini 1.5 & 0.0275 & 0.0175 & 0.0225 & 0.0167 & 0.0217 & 0.0175 \\
        Gemma 27B & 0.0242 & 0.0184 & 0.0275 & 0.0250 & 0.0259 & 0.0317 \\
        Gemma 9B & 0.0142 & 0.0150 & 0.0159 & 0.0142 & 0.0209 & 0.0209 \\
        LLaMA 70B & 0.0250 & 0.0376 & 0.0359 & 0.0250 & 0.0317 & 0.0342 \\
        LLaMA 8B & 0.0401 & 0.0376 & 0.0551 & 0.0351 & 0.0593 & 0.0426 \\
        Phi-3 Medium & 0.0234 & 0.0217 & 0.0167 & 0.0626 & 0.0175 & 0.0234 \\
        Phi-4 & 0.0142 & 0.0175 & 0.0159 & 0.0242 & 0.0309 & 0.0200 \\
        Mistral Large & 0.0267 & 0.0167 & 0.0417 & 0.0259 & 0.0459 & 0.0326 \\
        Mistral Saba & 0.0033 & 0.0134 & 0.0025 & 0.0083 & 0.0117 & 0.0117 \\
        ChatGPT-4o & 0.0067 & 0.0184 & 0.0200 & 0.0075 & 0.0109 & 0.0142 \\
        Qwen 32B & 0.0142 & 0.0109 & 0.0267 & 0.0250 & 0.0134 & 0.0192 \\
        \bottomrule
    \end{tabular}
    \caption{Performance Gap Analysis Across Datasets}
    \label{tab:4}
\end{table}

The Table~\ref{tab:4}  explains the difference in performance (accuracy) of all the 14 models across all the datasets. These gaps are explained in detail in  Section ~\ref{4.1}. The values in the table are presented in decimal form, whereas in the main paper Figure~\ref{fig:4}, they have been converted to percentages for ease of interpretation.

\begin{table}[H]
    \centering
    \renewcommand{\arraystretch}{1.2} 
    \begin{tabular}{lcccccc}
        \toprule
        & \textbf{Hti} & \textbf{Mld} & \textbf{Pak} & \textbf{Sol} & \textbf{Som} & \textbf{Sur} \\
        \midrule
        \textbf{Count} & 14 & 14 & 14 & 14 & 14 & 14 \\
        \textbf{Mean} & 0.0180 & 0.0192 & 0.0245 & 0.0234 & 0.0248 & 0.0231 \\
        \textbf{Std} & 0.0105 & 0.0090 & 0.0133 & 0.0141 & 0.0141 & 0.0096 \\
        \textbf{Min} & 0.0025 & 0.0042 & 0.0025 & 0.0075 & 0.0083 & 0.0067 \\
        \textbf{25\%} & 0.0123 & 0.0154 & 0.0161 & 0.0148 & 0.0144 & 0.0179 \\
        \textbf{50\% (Median)} & 0.0163 & 0.0179 & 0.0225 & 0.0246 & 0.0217 & 0.0221 \\
        \textbf{75\%} & 0.0248 & 0.0205 & 0.0288 & 0.0257 & 0.0301 & 0.0298 \\
        \textbf{Max} & 0.0401 & 0.0376 & 0.0551 & 0.0626 & 0.0593 & 0.0426 \\
        \bottomrule
    \end{tabular}
    \caption{Descriptive Statistics of accuracy drops across  models}
    \label{tab:5}
\end{table}

Table~\ref{tab:5} presents descriptive statistics of accuracy drops across different models for six datasets: Haiti (Hti), Moldova (Mld), Pakistan (Pak), Solomon Islands (Sol), Somalia (Som), and Suriname (Sur). The values represent the magnitude of performance drops when comparing each model’s accuracy on the culturally adapted datasets against the original GSM8K dataset. Count shows that all datasets have results from 14 different models. Min and Max indicate the smallest and largest observed accuracy drops, respectively. Percentiles (25\%, 50\% (Median), and 75\%) show the spread of accuracy drops.

\subsection{Difference in Tokenization}
\label{appendix:difference-in-tokenization}

\begin{figure}[H]
    \centering
    \includegraphics[width=1\textwidth]{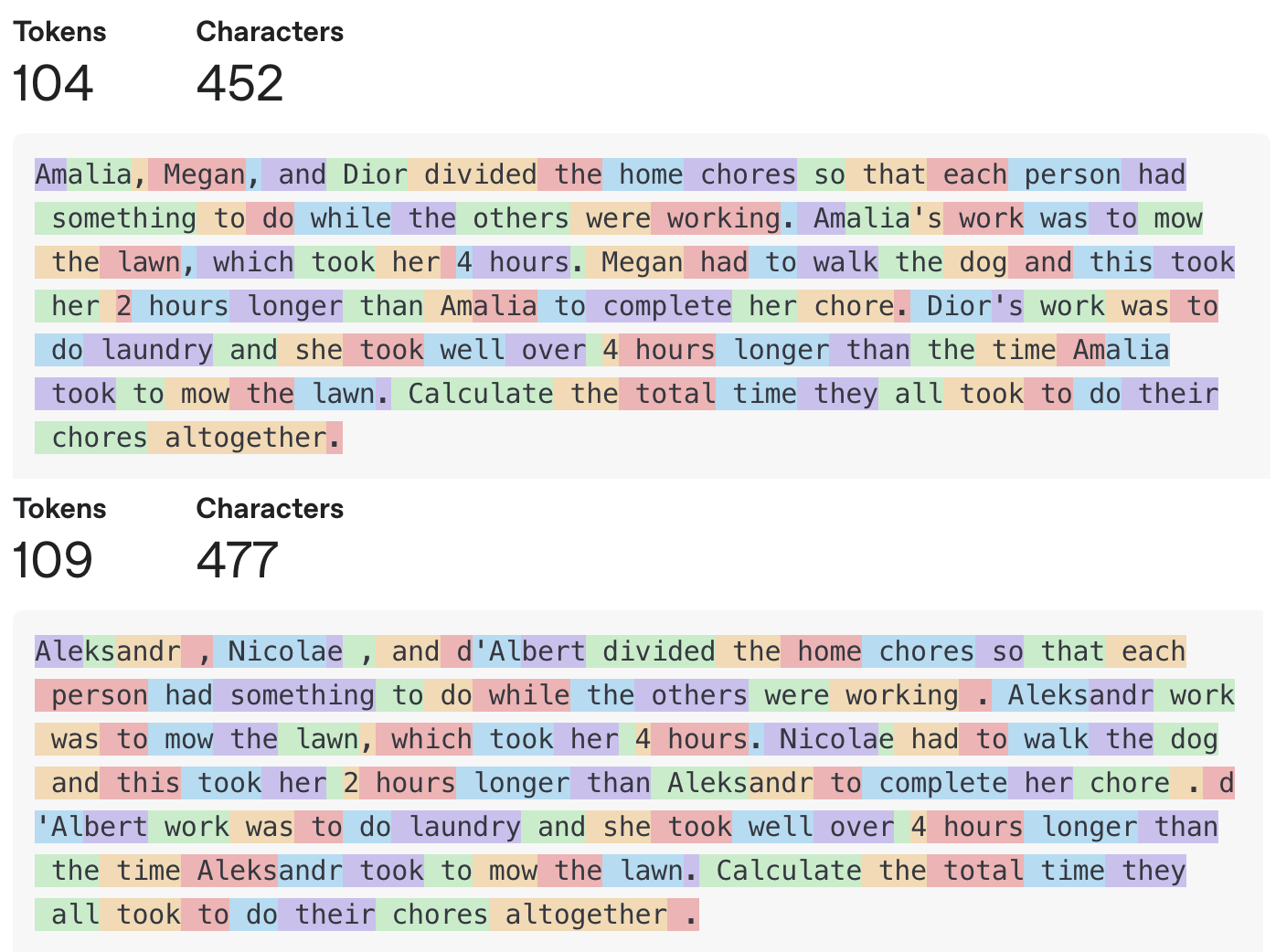}
    \caption{Difference in Tokenization}
    \label{fig:9}
\end{figure}

We utilize the OpenAI tokenizer to analyze how a question from the original GSM8K dataset is tokenized compared to its Moldovan-contextualized version. Notably, the only modification in this example is the change of names, yet the tokenization process treats them differently. \textbf{This discrepancy highlights how even minor cultural adaptations can alter tokenization, suggesting that culturally adapted vocabulary may be processed differently by the model, though the direct impact on reasoning performance cannot be established from tokenization differences alone.}

\subsection{McNemar Test}
\label{appendix:mcnemar-test}
\begin{table}[h]
    \centering
    \footnotesize
    \renewcommand{\arraystretch}{1.1}
    \setlength{\tabcolsep}{3pt}
    \begin{tabular}{lcccccc}
        \toprule
        \textbf{Model} & \textbf{Hti} & \textbf{Mld} & \textbf{Pak} & \textbf{Sol} & \textbf{Som} & \textbf{Sur} \\
        \midrule
        Mistral Saba  & 0.74933 & 0.12929 & 0.83585 & 0.36820 & 0.19335 & 0.17498 \\
                      & (46,42) & (57,41) & (48,45) & (55,45) & (57,43) & (53,39) \\
        \midrule
        Gem Flash 1.5-8B  & 0.00293$^{***}$ & 0.06171$^{*}$ & 0.01773$^{**}$ & 0.08241$^{*}$ & 0.01988$^{**}$ & 0.06399$^{*}$ \\
                          & (75,42) & (68,47) & (74,47) & (70,50) & (71,45) & (69,48) \\
        \midrule
        Gemma 2-27B  & 0.00346$^{***}$ & 0.01832$^{**}$ & 0.00119$^{***}$ & 0.00288$^{***}$ & 0.00152$^{***}$ & 0.00021$^{***}$ \\
                     & (61,32) & (51,29) & (66,33) & (63,33) & (61,30) & (70,32) \\
        \midrule
        LLaMA 3.1-70B  & 0.00231$^{***}$ & 0.00001$^{***}$ & 0.00003$^{***}$ & 0.00423$^{***}$ & 0.00018$^{***}$ & 0.00007$^{***}$ \\
                       & (61,31) & (75,30) & (73,30) & (67,37) & (69,31) & (72,31) \\
        \midrule
        Gemma 2-9B  & 0.16826 & 0.10461 & 0.10420 & 0.15212 & 0.04588$^{**}$ & 0.04140$^{**}$ \\
                    & (76,59) & (64,46) & (71,52) & (71,54) & (85,60) & (82,57) \\
        \midrule
        Phi-4  & 0.06037$^{*}$ & 0.02203$^{**}$ & 0.05025$^{*}$ & 0.00169$^{***}$ & 0.00016$^{***}$ & 0.00631$^{***}$ \\
               & (45,28) & (49,28) & (52,33) & (55,26) & (65,28) & (48,24) \\
        \midrule
        DeepSeek  & 0.14564 & 0.00804$^{***}$ & 0.00280$^{***}$ & 0.00016$^{***}$ & 0.00734$^{***}$ & 0.00169$^{***}$ \\
                  & (47,33) & (54,29) & (52,25) & (62,26) & (57,31) & (55,26) \\
        \midrule
        Gem Flash 2.0  & 0.00094$^{***}$ & 0.00061$^{***}$ & 0.00000$^{***}$ & 0.00027$^{***}$ & 0.00000$^{***}$ & 0.00002$^{***}$ \\
                        & (32,10) & (33,10) & (46,11) & (33,9) & (41,8) & (38,9) \\
        \midrule
        Phi-3 Medium  & 0.08496$^{*}$ & 0.10346 & 0.20416 & 0.00000$^{***}$ & 0.18424 & 0.07479$^{*}$ \\
                      & (137,109) & (131,105) & (122,102) & (161,86) & (124,103) & (129,101) \\
        \midrule
        Mistral Large  & 0.00031$^{***}$ & 0.03079$^{**}$ & 0.00000$^{***}$ & 0.00117$^{***}$ & 0.00000$^{***}$ & 0.00002$^{***}$ \\
                        & (54,22) & (49,29) & (66,16) & (59,28) & (75,20) & (61,22) \\
        \midrule
        GPT-4o  & 0.33175 & 0.00535$^{***}$ & 0.00427$^{***}$ & 0.27168 & 0.11116 & 0.02701$^{**}$ \\
                     & (30,22) & (40,18) & (45,21) & (31,22) & (35,22) & (35,18) \\
        \midrule
        Qwen 2.5-32B  & 0.06755$^{*}$ & 0.19276 & 0.00111$^{***}$ & 0.00161$^{***}$ & 0.10523 & 0.02202$^{**}$ \\
                       & (47,30) & (49,36) & (62,30) & (58,28) & (51,35) & (58,35) \\
        \midrule
        Claude 3.5  & 0.74283 & 0.47313 & 0.06599$^{*}$ & 0.09874$^{*}$ & 0.14331 & 0.24298 \\
                     & (20,17) & (18,13) & (28,15) & (20,10) & (24,14) & (22,14) \\
        \midrule
        LLaMA 3.1-8B  & 0.00674$^{***}$ & 0.00879$^{***}$ & 0.00017$^{***}$ & 0.01628$^{**}$ & 0.00005$^{***}$ & 0.00242$^{***}$ \\
                      & (175,127) & (164,119) & (184,118) & (167,125) & (185,114) & (162,111) \\
        \bottomrule
    \end{tabular}
    \caption{McNemar Test Results for Model Performance Across Datasets. Values represent p-values (rounded to 5 decimal places). Significance: $^{*} p<0.10$, $^{**} p<0.05$, $^{***} p<0.01$. (b,c) values in parentheses.}
    \label{tab:6}
\end{table}

The McNemar test results provide a statistical measure of whether the accuracy of different models significantly dropped when tested on culturally adapted datasets. This helps us understand which models are more sensitive to cultural shifts and which ones are more robust. Each cell in the table contains a p-value, along with two numbers (b, c) in parentheses.
\pagebreak
\begin{itemize}
    \item \textbf{b} represents the number of times the model got the question right on the original GSM8K dataset but wrong on the culturally adapted dataset. \textbf{c} represents the number of times the model got the question wrong on GSM8K but right on the adapted dataset. A higher \textbf{b} compared to \textbf{c} indicates that the model struggles more with the adapted dataset.
    \item A low p-value (p $<$ 0.05) means the accuracy drop is statistically significant and not due to random chance.

\end{itemize}

Certain models showed significant performance drops, suggesting that they struggle to generalize to culturally adapted datasets.
\begin{itemize}
    \item \textbf{LLaMA 3.1-70B} consistently had high b-values across all datasets, meaning it frequently failed on culturally adapted versions of the questions while performing well on the original GSM8K. It had some of the lowest p-values (p $<$ 0.01), particularly in Moldova, Pakistan, Somalia, and Suriname. This suggests that the model’s pretraining data might not be diverse enough to handle different cultural contexts in mathematical reasoning.
    \item \textbf{Gemini Flash 2.0} also exhibited significant accuracy drops on all datasets, with particularly large b-values in Pakistan, Solomon Islands, and Somalia. The very low p-values indicate that these failures were systematic rather than random. This suggests that Gemini Flash 2.0 may have a strong Western-centric bias, causing difficulties in understanding culturally adapted variations of math problems.
    \item \textbf{Mistral Large 2411} aced similar issues, with substantial accuracy drops across all datasets, particularly in Pakistan and Somalia. The fact that b-values are consistently high compared to c-values means the model performs well in the original setting but fails when cultural factors are introduced.    
\end{itemize}
Some models had moderate but still statistically significant performance drops, meaning they weren’t completely failing, but they still showed weaknesses.
\begin{itemize}
    \item \textbf{DeepSeek} showed a notable accuracy drop in Moldova, Pakistan, Solomon Islands, and Somalia. The p-values are under 0.01 in these cases, showing that model probably  struggles with mathematical reasoning when cultural adaptations are introduced. However, its b-values are not as extreme as those of Gemini Flash or LLaMA 3.1-70B, suggesting some ability to adapt.
    \item \textbf{Phi-4} also showed significant drops in Moldova, Solomon Islands, and Somalia, with p-values below 0.05. While the drop was not as severe as in LLaMA or Gemini Flash, it suggests that Phi-4 might not generalize well to unfamiliar cultural settings in solving math problems.
     \item \textbf{Gemma 2-27B} showed consistent accuracy reductions across all datasets, particularly in Pakistan, Somalia, and Suriname. The p-values and high b-values confirm that the model is sensitive to cultural variations.    
\end{itemize}

The McNemar test results show that certain models, like LLaMA 3.1-70B and Gemini Flash 2.0, struggle significantly with cultural variations, while others, like Mistral Saba and Claude 3.5, remain more stable.

\subsection{Qualitative Error Analysis}
\label{appendix:qualitative-error-analysis}

Qualitative error analysis examines mathematical reasoning errors in culturally adapted versions of GSM8K problems across 14 models, each evaluated over three runs. While the mathematical logic remains unchanged, cultural adaptations of GSM8K introduce linguistic and contextual variations. We assess correctness by comparing model-generated answers to the GSM8K ground truth, identifying discrepancies caused by reasoning errors or cultural misinterpretations. The goal is to uncover error patterns and model weaknesses, determining whether failures are universally difficult or model-specific.

A question is marked incorrect if at least one of the three runs produces an incorrect reasoning answer, rather than relying on majority voting. This approach captures model inconsistencies, highlighting cases where failures occur occasionally rather than systematically. Each answer is compared against the GSM8K ground truth to distinguish genuine numerical reasoning errors from those influenced by cultural adaptations. By flagging errors based on a single incorrect run, we account for response variability, identifying unstable performance patterns that might be overlooked with a stricter majority-based approach.

\subsubsection{Currency-Based Errors}
\label{appendix:currency-based-errors}

One of the key errors we identify is the models' inconsistency in handling mathematical reasoning across different currency units. For example, When solving a problem with a cost of 0.1 USD per cubic foot, almost all models correctly compute the total cost. However, when the same problem uses another cultures currency units such as, 0.1 HTG (Haitian Gourde) per cubic foot, the models incorrectly return an answer as if the cost is 1 HTG per cubic foot, leading to a tenfold overestimation.

\begin{figure}[H]
    \centering
    \includegraphics[width=1\textwidth]{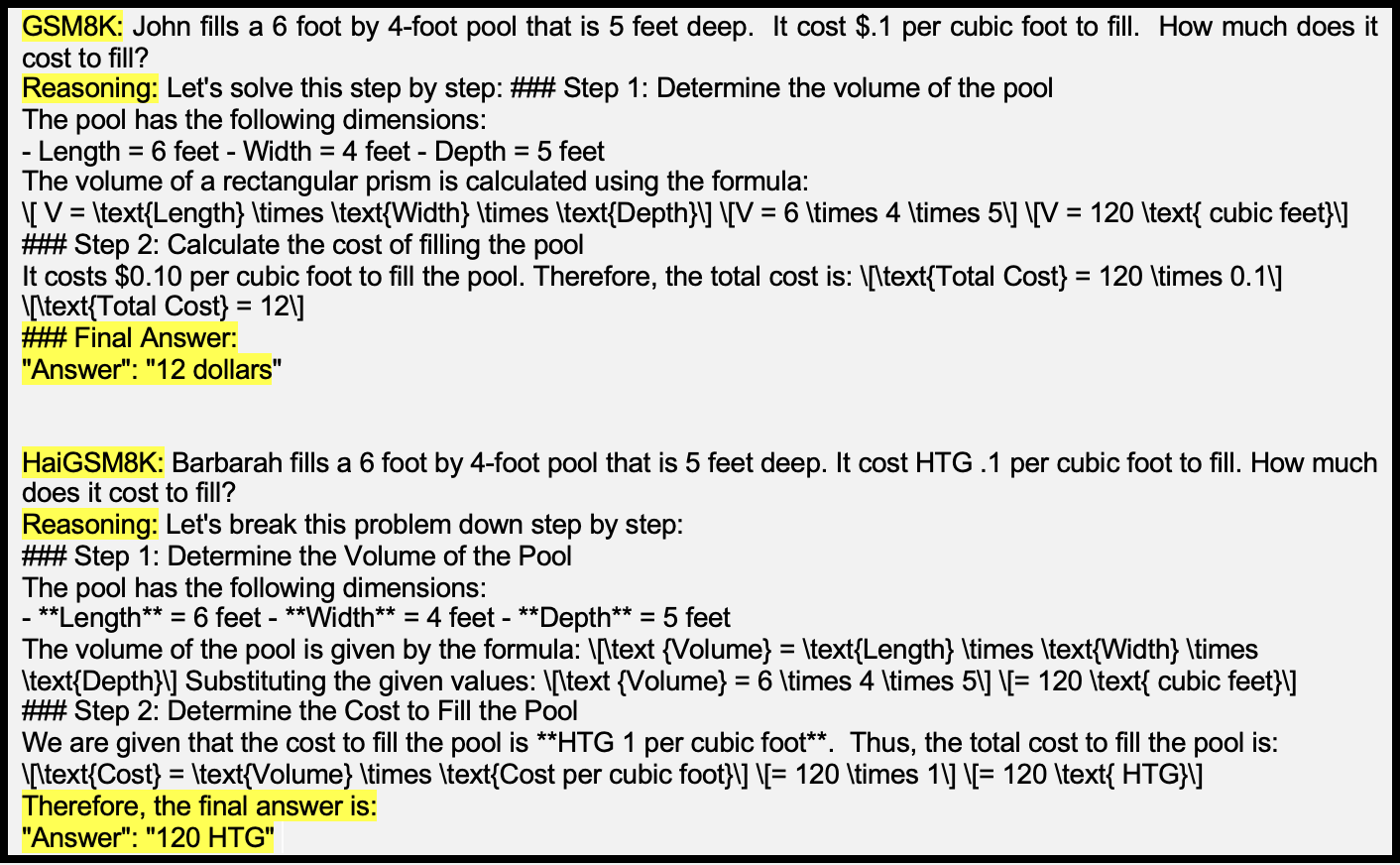}
    \caption{GPT-4o Reasoning}
    \label{fig:10}
\end{figure}

This discrepancy suggests a contextual bias in numerical interpretation—while the model accurately processes decimals in USD, it defaults to whole-number assumptions for HTG. This aligns with real-world usage, as HTG is rarely used in decimal form due to inflation and economic practices, with transactions typically rounded to whole numbers. However, this exposes a bias in numerical reasoning, where the model fails to apply consistent mathematical principles across currencies and instead relies on learned heuristics from their training data. Additionally, the models reasoning process differed based on currency—when using USD, they explicitly recognize and multiply by 0.1, but with HTG, they misinterpret the decimal, either skipping or rounding the value incorrectly. This highlights a deeper issue: models do not merely struggle with arithmetic but also with the contextual expectations of numerical formats in different cultures, leading to systematic reasoning errors.

\subsubsection{Errors in Cultural Entity Interpretation}
\label{appendix:errors-in-cultural-entity-interpretation}
Another key error pattern observed in our qualitative analysis is the models inconsistent handling of mathematical reasoning when a question is culturally adapted. For example, in the original GSM8K dataset, the question states: `Tom decides to give his wife an anniversary getaway.' The models correctly identifies that two plane tickets are required and accurately compute the total trip cost. However, when the question is culturally adapted—such as `Mary decides to give her tambu man (father-in-law) a Christmas getaway' (Solomon Islands context) or `Khalid Sulehri decides to give his Jija (brother-in-law) an Eid-ul-Fitar getaway' (Pakistani context)—the models incorrectly assumes only a single traveler, leading to errors in the total cost calculation.

\begin{figure}[H]
    \centering
    \includegraphics[width=1\textwidth]{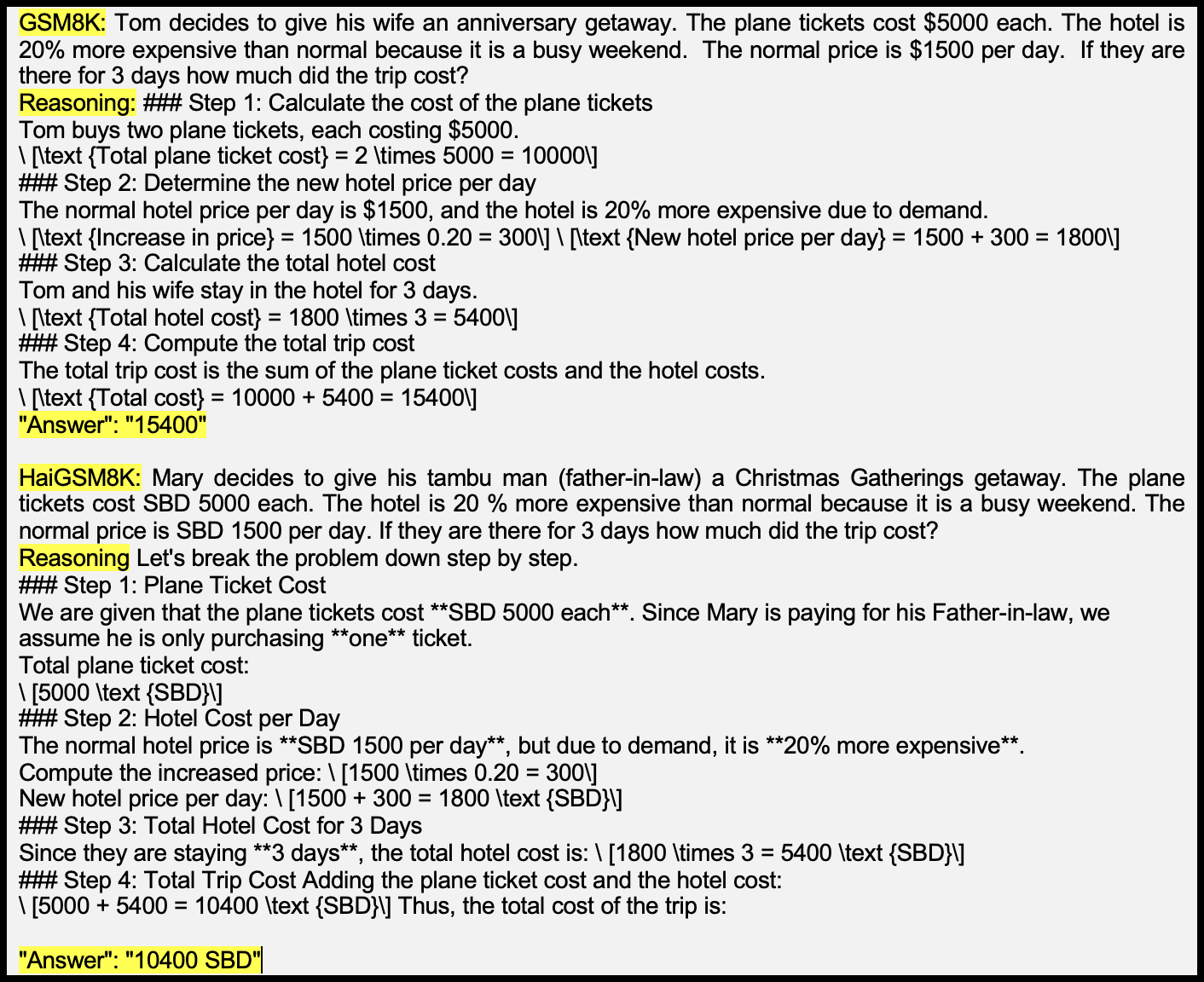}
    \caption{GPT-4o Reasoning}
    \label{fig:11}
\end{figure}

This pattern suggests a mathematical reasoning error when interpreting different cultural entities. While the models correctly process common familial relationships (husband-wife), it struggles with non-Western family structures (tambu man, Jija). Even though the numerical reasoning should remain unchanged, the errors indicate implicit biases in how the models associates relationships with travel expectations, affecting their ability to generalize reasoning across cultures.

\subsubsection{Problem Interpretation Errors}

We also identify a contextual misinterpretation error that lead to a systematic inaccuracies. The original question, in GSM8K, states: `A pet store currently has 5 dogs, 2 cats, and 10 birds. How many legs in total do the pets in the store have?' All models tested produced the correct response when presented with this version. However, when the question is culturally adapted to, lets say Somalian culture, replacing `dog,' `cat,' and `bird' with `maroodi,' `shabeel,' and `gorgor,' the models consistently produce incorrect answers. 

\begin{figure}[H]
    \centering
    \includegraphics[width=1\textwidth]{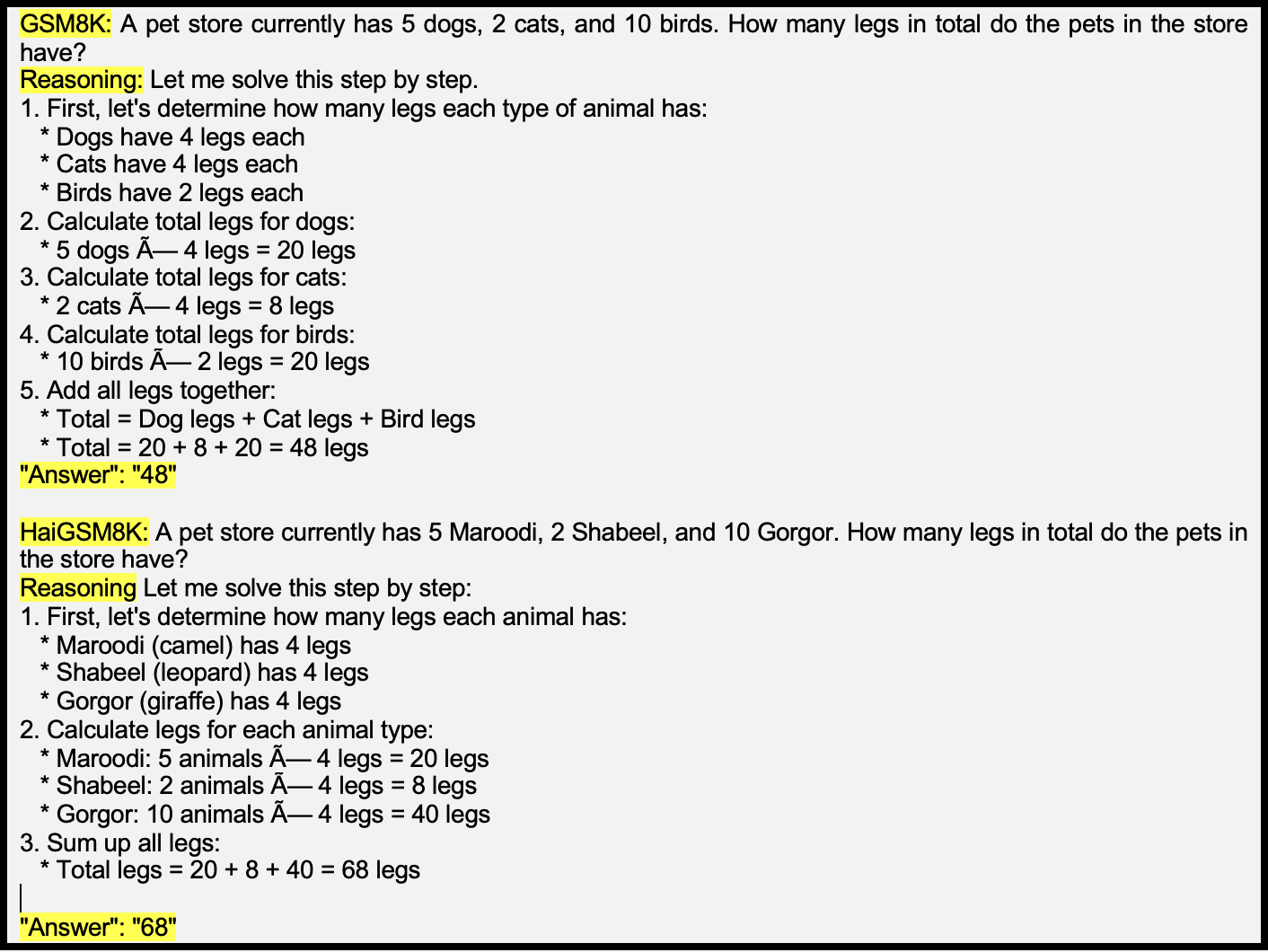}
    \caption{Claud-3.5-sonnet Reasoning}
    \label{fig:12}
\end{figure}

This finding highlights a contextual misinterpretation error, where models fail to correctly associate culturally specific terms with their actual meanings. Rather than understanding the new entities within their cultural and linguistic context, the models defaulted to familiar patterns from their training data. 

\subsubsection{Mathematical Reasoning Errors}
Across our evaluation, we observed mathematical reasoning errors where cultural adaptation caused models to fundamentally alter their solution strategy despite the underlying logic remaining unchanged. A representative example involves a gift bag planning problem where the 0.75 bags-per-guest ratio already accounts for the 1/4 of guests who do not show up. In the original GSM8K version, the model correctly applies this ratio directly to the 16 invited guests, computing 0.75 × 16 = 12 bags and arriving at a total cost of \$24. However, in the Surinamese adaptation, the model introduces a redundant step, first computing expected attendees by applying the 1/4 no-show rate to arrive at 12, and then applying the 0.75 ratio again to this already-reduced figure, producing 9 bags and a total cost of SRD 18. The same mathematical structure that was handled correctly in the original framing breaks down once the cultural context shifts.

\begin{figure}[H]
    \centering
    \includegraphics[width=1\textwidth]{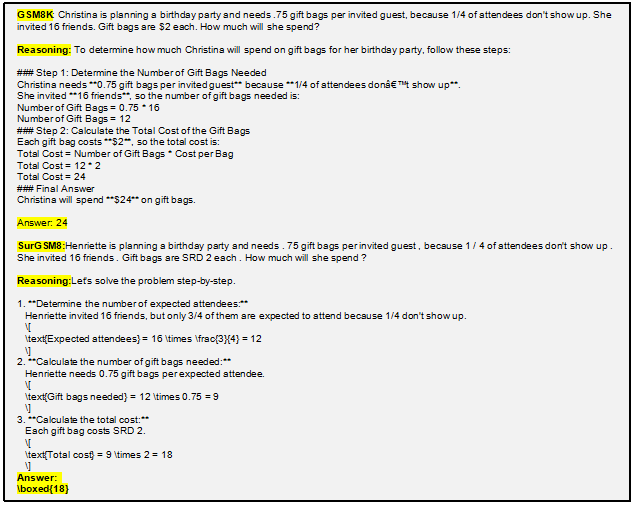}
    \caption{Claud-3.5-sonnet Reasoning}
    \label{fig:12}
\end{figure}
\vspace{0.5cm}

\subsubsection{Calculation Errors}
Unlike reasoning errors where the solution strategy itself breaks down, the calculation errors we observed involve models that correctly structured their approach but produced wrong final answers due to arithmetic mistakes introduced after cultural adaptation. A representative example involves a week-long attendance counting problem where the original GSM8K version is solved correctly, with the model accurately summing 50, 70, 140, and 240 people across the seven days to arrive at 500. In the Somalian adaptation, replacing the character name and substituting ``beach'' with ``Tuulo'', the model correctly computes the count for each individual day yet arrives at an incorrect total of 400, omitting the day one count from the final summation. The same arithmetic, performed correctly in the original framing, breaks down once the cultural context shifts.

\begin{figure}[H]
    \centering
    \includegraphics[width=1\textwidth]{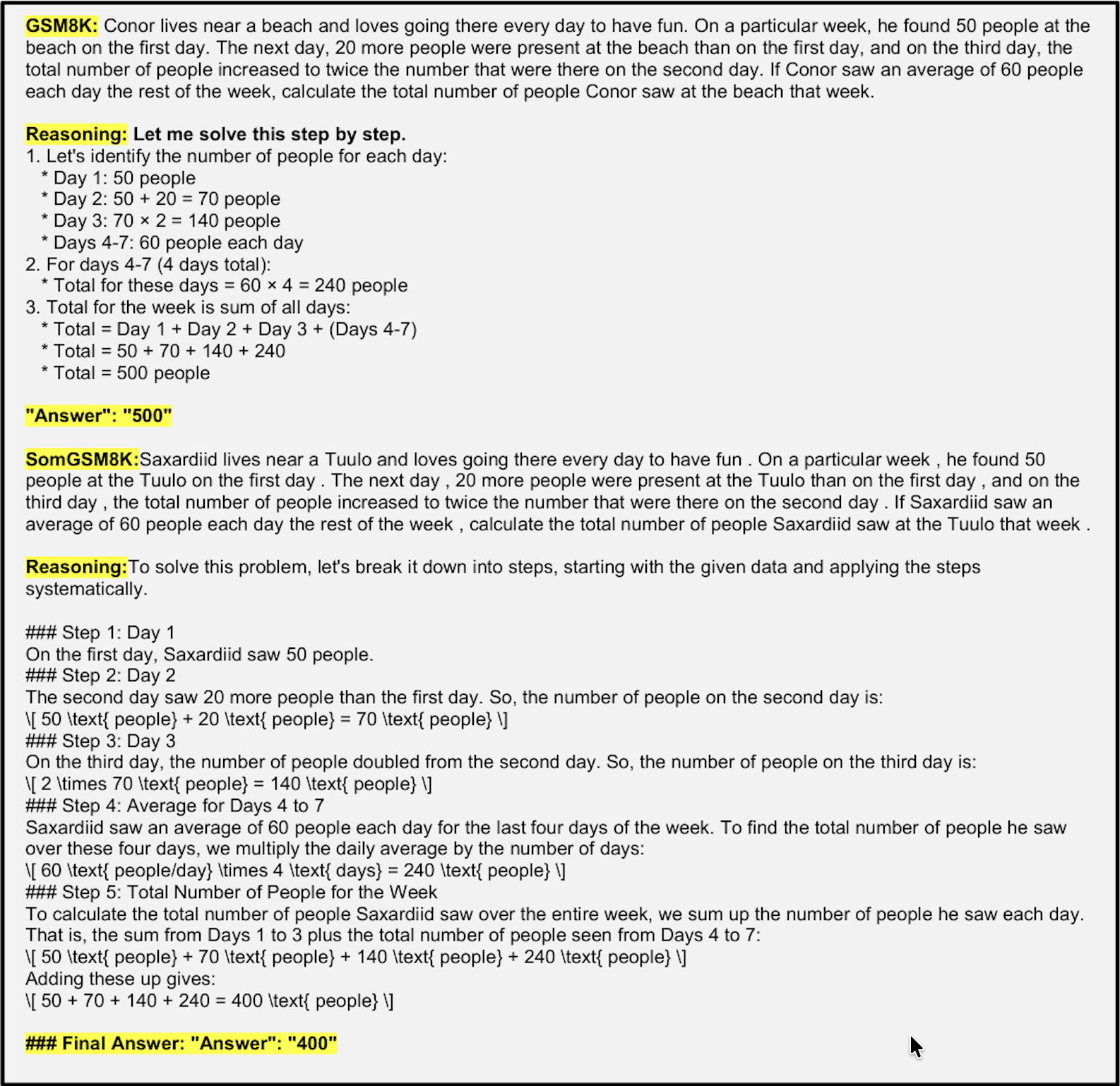}
    \caption{Qwen 2.5-32B-Instruct reasoning}
    \label{fig:12}
\end{figure}

\subsubsection{Other / unclear errors}
A small set of errors in our analysis cannot be attributed to mathematical reasoning or calculation failures, and instead arise from issues such as answer extraction mismatches, formatting inconsistencies, and non-numeric outputs. For example, in a Moldovan-adapted history test scoring problem, the model correctly calculates 8 points from multiple-choice questions, 18 from true/false questions, and 15 from long-answer questions, explicitly stating a correct total of 41 points within its reasoning trace, yet the final answer field returns 3, the number of correctly answered long-answer questions rather than the total score.

\begin{figure}[H]
    \centering
    \includegraphics[width=0.8\textwidth]{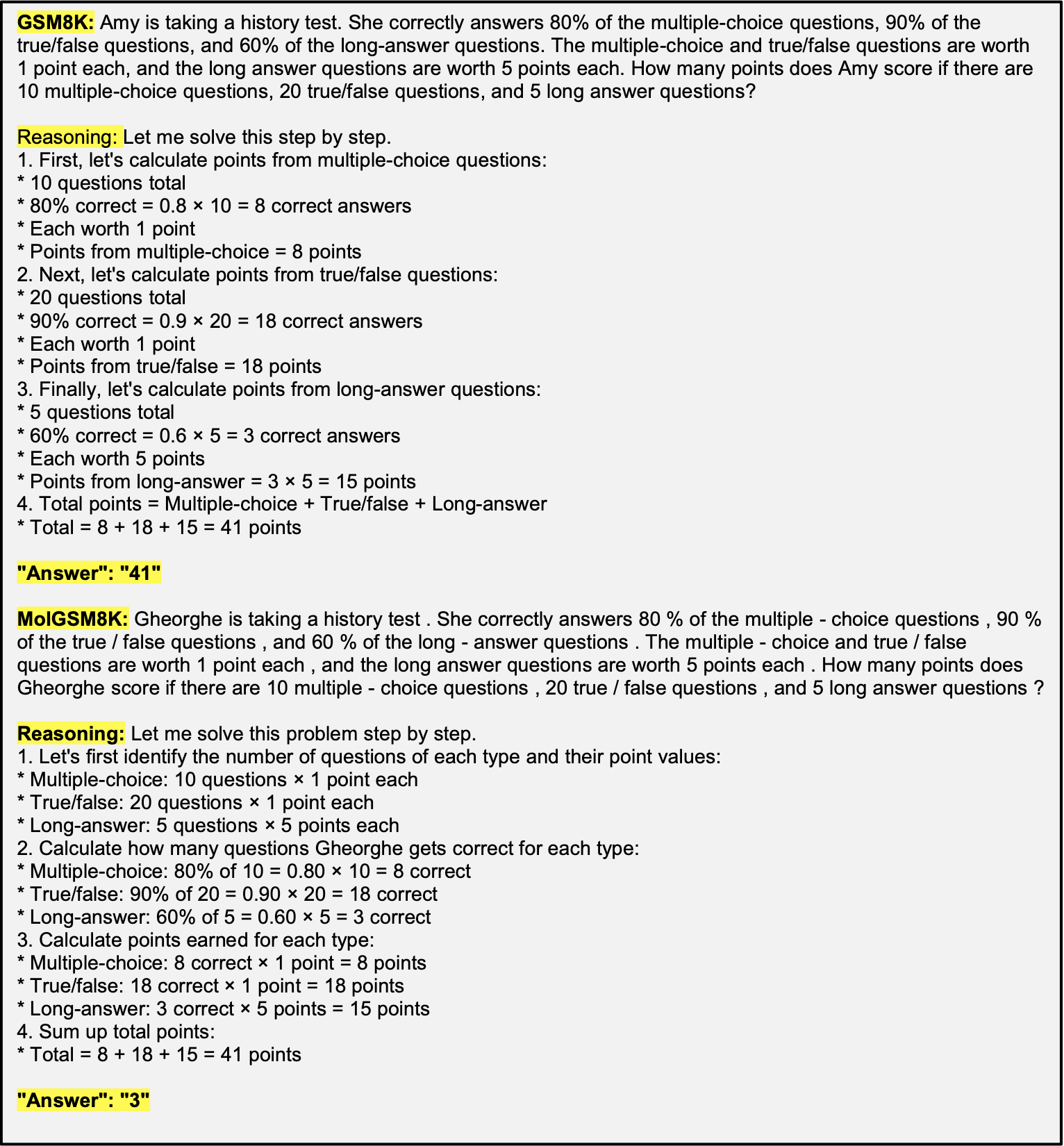}
    \caption{Claud-3.5-sonnet Reasoning}
    \label{fig:12}
\end{figure}

\subsection{Quantitative Error Analysis}
\label{quantitative-error-analysis}
To systematically quantify error patterns across all models and cultural variants, we conducted a comprehensive annotation of all incorrect model responses identified in our evaluation. We collected all cases where at least one of the three model runs produced an incorrect answer according to the strict consistency criterion described in Section ~\ref{4.1}. This resulted in 18,887 error instances across 14 models and 7 datasets (original GSM8K plus 6 cultural variants).

We removed cases involving only minor rounding or formatting differences that did not reflect genuine reasoning errors. Duplicate entries for the same model-culture-question combination were also removed to ensure each unique error was counted only once. Each error was categorized into one of six types: (1) Calculation error, (2) Mathematical Reasoning error, (3) Relationship misunderstanding, (4) Unit/currency error, (5) Problem misinterpretation, and (6) Other/unclear. The categorization was performed using an LLM-assisted annotation script that analyzed the model's reasoning process for the first failing run among the three attempts.

To ensure annotation quality, we manually reviewed 50 randomly sampled error cases. The manual review confirmed that the LLM-based categorization was largely accurate, with minor discrepancies primarily occurring in the "Other/unclear" category, which represents only 0.76\% of total errors. 

\subsubsection{Error Distribution by Culture}

\begin{table}[H]
\centering
\footnotesize
\renewcommand{\arraystretch}{1.1}
\setlength{\tabcolsep}{3pt}
\begin{tabular}{lcccccccc}
\toprule
\textbf{Error Category} & \textbf{GSM8K} & \textbf{Hti} & \textbf{Mld} & \textbf{Pak} & \textbf{Sol} & \textbf{Som} & \textbf{Sur} \\
\midrule
Mathematiical Reasoning error & 1362 & 1470 & 1475 & 1490 & 1525 & 1508 & 1501 \\
                        & (58.58\%) & (54.63\%) & (54.61\%) & (53.20\%) & (54.78\%) & (53.49\%) & (54.27\%) \\
\midrule
Calculation error & 890 & 918 & 893 & 1002 & 881 & 1007 & 924 \\
                  & (38.62\%) & (34.11\%) & (33.06\%) & (35.77\%) & (31.65\%) & (35.72\%) & (33.41\%) \\
\midrule
Relationship misunderstanding & 65 & 119 & 140 & 103 & 195 & 85 & 144 \\
                              & (2.80\%) & (4.42\%) & (5.18\%) & (3.68\%) & (7.00\%) & (3.02\%) & (5.21\%) \\
\midrule
Unit / currency error & 0 & 137 & 129 & 113 & 97 & 146 & 124 \\
                      & (0\%) & (5.09\%) & (4.78\%) & (4.03\%) & (3.48\%) & (5.18\%) & (4.48\%) \\
\midrule
Problem misinterpretation & 0 & 23 & 47 & 65 & 66 & 40 & 51 \\
                          & (0\%) & (0.85\%) & (1.74\%) & (2.32\%) & (2.37\%) & (1.42\%) & (1.84\%) \\
\midrule
Other / unclear & 0 & 24 & 17 & 28 & 20 & 33 & 22 \\
                & (0\%) & (0.89\%) & (0.63\%) & (1.00\%) & (0.72\%) & (1.17\%) & (0.80\%) \\
\bottomrule
\end{tabular}
\caption{Error distribution by cultural variant. Values show count with percentage in parentheses. Hti = HaiGSM8K, Mld = MolGSM8K, Pak = PakGSM8K, Sol = SolIGSM8K, Som = SomGSM8K, Sur = SurGSM8K.}
\label{tab:error_by_culture}
\end{table}

Table ~\ref{tab:error_by_culture} reveals notable variation in culturally-specific error patterns across datasets. Solomon Islands exhibits the highest proportion of relationship misunderstandings (7.0\%), consistent with the use of unfamiliar kinship terms such as "tambu man." Somalia and Haiti show elevated unit/currency errors (5.18\% and 5.09\%, respectively), reflecting challenges with less common currency units like the Somali Shilling and Haitian Gourde. The original GSM8K dataset shows the lowest rates of these culturally-specific errors, as expected.

\subsubsection{Error Distribution by Model}
\begin{table}[H]
\centering
\footnotesize
\renewcommand{\arraystretch}{1.1}
\setlength{\tabcolsep}{3pt}
\begin{tabular}{lcccccc}
\toprule
\textbf{Model} & \textbf{Calculation} & \textbf{Reason} & \textbf{Relation} & \textbf{Currency} & \textbf{Misinterp} & \textbf{Other/Unclear} \\
\midrule
Claude 3.5  & 185 & 163 & 13 & 36 & 13 & 4 \\
     & (44.69\%) & (39.37\%) & (3.14\%) & (8.70\%) & (3.14\%) & (0.97\%) \\
\midrule
DeepSeek  & 285 & 305 & 55 & 52 & 20 & 9 \\
      & (39.26\%) & (42.01\%) & (7.58\%) & (7.16\%) & (2.75\%) & (1.24\%) \\
\midrule
Gem Flash 2.0  & 214 & 325 & 37 & 23 & 20 & 5 \\
     & (34.29\%) & (52.08\%) & (5.93\%) & (3.69\%) & (3.21\%) & (0.80\%) \\
\midrule
Gem Flash 1.5-8B & 507 & 883 & 54 & 61 & 22 & 8 \\
     & (33.03\%) & (57.52\%) & (3.52\%) & (3.97\%) & (1.43\%) & (0.52\%) \\
\midrule
Gemma 2-27B & 538 & 620 & 29 & 50 & 20 & 10 \\
     & (42.46\%) & (48.93\%) & (2.29\%) & (3.95\%) & (1.58\%) & (0.79\%) \\
\midrule
Gemma 2-9B & 686 & 775 & 51 & 56 & 17 & 3 \\
     & (43.20\%) & (48.80\%) & (3.21\%) & (3.53\%) & (1.07\%) & (0.19\%) \\
\midrule
LLaMA 3.1-70B & 848 & 2211 & 117 & 68 & 20 & 12 \\
     & (25.89\%) & (67.49\%) & (3.57\%) & (2.08\%) & (0.61\%) & (0.37\%) \\
\midrule
LLaMA 3.1-8B & 840 & 2217 & 110 & 68 & 23 & 14 \\
     & (25.67\%) & (67.76\%) & (3.36\%) & (2.08\%) & (0.70\%) & (0.43\%) \\
\midrule
Phi-3 Medium & 780 & 1040 & 90 & 63 & 37 & 39 \\
     & (38.07\%) & (50.76\%) & (4.39\%) & (3.07\%) & (1.81\%) & (1.90\%) \\
\midrule
Phi-4 & 345 & 336 & 84 & 54 & 22 & 3 \\
     & (40.88\%) & (39.81\%) & (9.95\%) & (6.40\%) & (2.61\%) & (0.36\%) \\
\midrule
Mistral Large & 299 & 365 & 52 & 50 & 20 & 19 \\
      & (37.14\%) & (45.34\%) & (6.46\%) & (6.21\%) & (2.48\%) & (2.36\%) \\
\midrule
Mistral Saba  & 436 & 465 & 65 & 59 & 15 & 2 \\
      & (41.84\%) & (44.63\%) & (6.24\%) & (5.66\%) & (1.44\%) & (0.19\%) \\
\midrule
GPT-4o & 266 & 214 & 43 & 47 & 19 & 7 \\
     & (44.63\%) & (35.91\%) & (7.21\%) & (7.89\%) & (3.19\%) & (1.17\%) \\
\midrule
Qwen 2.5-32B  & 294 & 412 & 51 & 59 & 24 & 9 \\
     & (34.63\%) & (48.53\%) & (6.01\%) & (6.95\%) & (2.83\%) & (1.06\%) \\
\bottomrule
\end{tabular}
\caption{Error distribution by model. Reason = Mathematical reasoning error, Relation = Relationship misunderstanding, Currency = Unit/currency error, Misinterp = Problem misinterpretation.}
\label{tab:error_by_model}
\end{table}
Table ~\ref{tab:error_by_model} shows that error distribution patterns vary across models. Claude 3.5 Sonnet (C3.5) exhibits a notably higher proportion of calculation errors (44.69\%) compared to reasoning errors (39.37\%), contrasting with most other models where reasoning errors dominate. Claude 3.5 also shows the highest rate of unit/currency errors (8.70\%), suggesting sensitivity to unfamiliar numerical contexts despite overall strong performance. Smaller models like LLaMA 3.1-8B show the highest absolute error counts, consistent with their lower overall accuracy reported in Section 4.2.2.  Across all models, reasoning and calculation errors account for over 80\% of failures, with culturally-specific errors (relationship and currency) comprising a smaller but consistent proportion.

\subsubsection{Error Distribution by Entity Category}
\label{Error Distribution by Entity Category}

To complement the error type analysis in Section 12.1 and 12.2, Table \ref{tab:error_by_category} examines which cultural entity categories appear disproportionately more frequently in incorrect responses relative to their overall occurrence across the 1,198 culturally adaptable questions.

\begin{table}[H]
\centering
\renewcommand{\arraystretch}{1.5}
\begin{tabular}{>{\raggedright\arraybackslash}p{4.5cm}>{\centering\arraybackslash}p{2.5cm}>{\centering\arraybackslash}p{2cm}>{\centering\arraybackslash}p{2.5cm}}
\hline\hline
\textbf{Category} & \textbf{Occurrence \%} & \textbf{Error \%} & \textbf{Difference} \\
\hline\hline
Commerce and Economy & 35.2\% & 44.8\% & {+9.6\%} \\
Food and Drink       & 14.4\% & 22.2\% & {+7.8\%} \\
Other                & 3.6\%  & 4.1\%  & +0.5\% \\
Education            & 0.9\%  & 1.2\%  & +0.3\% \\
Places and Locations & 10.7\% & 10.9\% & +0.2\% \\
Clothing and Appearance & 1.6\% & 1.6\% & 0.0\% \\
People and Roles     & 90.7\% & 89.2\% & $-$1.5\% \\
Culture and Arts     & 5.6\%  & 4.1\%  & $-$1.5\% \\
\hline\hline
\end{tabular}
\caption{Error Distribution by Cultural Entity Category}
\label{tab:error_by_category}
\end{table}

The Difference column indicates whether a category appears disproportionately more or less in incorrect responses relative to its overall occurrence. A positive value indicates the category is associated with more errors than its frequency alone would predict. Commerce and Economy entities show the largest positive difference ($+$9.6\%), followed by Food and Drink ($+$7.8\%), consistent with the currency-related and cultural entity interpretation errors,  identified in the qualitative analysis in Section \ref{4.2.4}. In contrast, People and Roles shows a slightly negative difference ($-$1.5\%), indicating that despite person names appearing in over 90\% of questions, their substitution does not disproportionately cause errors. Note that this category is dominated by Person name entities (87.1\% of questions). This dominance  dilutes the signal from family member entities (4.9\%) that were identified as error-prone in the qualitative analysis.

\end{document}